%% file: main.tex
\documentclass{article}

\usepackage[main,final]{neurips_2026}
\setcitestyle{numbers,square,comma}

\usepackage[utf8]{inputenc}
\usepackage[T1]{fontenc}
\usepackage{microtype}
\usepackage{graphicx}
\usepackage{subcaption}
\usepackage{booktabs}
\usepackage{hyperref}
\usepackage{url}
\usepackage{amsmath}
\usepackage{amssymb}
\usepackage{amsfonts}
\usepackage{mathtools}
\usepackage{amsthm}
\usepackage[table]{xcolor}
\usepackage{multirow}
\usepackage{wrapfig}
\usepackage[capitalize,noabbrev]{cleveref}
\usepackage[textsize=tiny]{todonotes}
\usepackage{courier} 

\theoremstyle{plain}

\theoremstyle{definition}

\theoremstyle{remark}

\title{FAVLA: A Force-Adaptive Multi-Rate VLA Model for Contact-Rich Robotic Manipulation}

\author{%
\normalfont\bfseries Yao Li\textsuperscript{1,2*}\quad
  Peiyuan Tang\textsuperscript{3,4*}\quad
  Wuyang Zhang\textsuperscript{1,2,4\textdagger}\quad
  Haojie Ren\textsuperscript{5}\quad 
  Chengyang Zhu\textsuperscript{1}\quad  \\
\normalfont\bfseries  Yifan Duan\textsuperscript{5}\quad 
  Weikai Shi\textsuperscript{5}\quad 
  Xiaodong Zhang\textsuperscript{5}\quad 
  Yuming Dong\textsuperscript{6}\quad
  Zijiang Yang\textsuperscript{5}\quad \\
\normalfont\bfseries  Jianmin Ji\textsuperscript{5}\quad 
  Yanyong Zhang\textsuperscript{1\textdagger}\quad \\
  \parbox{0.96\textwidth}{\centering\normalfont\small
    \textsuperscript{1}School of Artificial Intelligence and Data Science, University of Science and Technology of China \\
    \textsuperscript{2}Suzhou Institute for Advanced Research, University of Science and Technology of China \\  
    \textsuperscript{3}School of Computer Science and Technology, Xi'an Jiaotong University \\
    \textsuperscript{4}Suzhou Artificial Intelligence Laboratory \\
    \textsuperscript{5}School of Computer Science and Technology, University of Science and Technology of China \\
    \textsuperscript{6}Institute of Intelligent Driving, Jianghuai Automobile Group Corp., Ltd \\
    \textsuperscript{*}Equal contribution\quad
    \textsuperscript{\textdagger}Corresponding authors
    }
}

\begin{document}

\maketitle

\input{sections/abstract}

\input{sections/introduction}

\input{sections/related_work}

\input{sections/methods}

\input{sections/Experiments}

\input{sections/Conclusion}

\begin{ack}
This work was supported by the New Generation Artificial Intelligence-National Science and Technology Major Project (Grant No.2025ZD0122600) and in part by the Jiangsu Funding Program for Excellent Postdoctoral Talent (Grant No.2025ZB315).
The authors declare no competing interests.
\end{ack}


\newpage
\bibliography{example_paper}
\bibliographystyle{plainnat}

\newpage
\appendix
\onecolumn

\input{sections/Appendix}

\newpage
\input{checklist.tex}

\end{document}

%% file: sections/abstract.tex
\begin{abstract}
Vision-Language-Action (VLA) models have shown strong potential for general robotic manipulation, but contact-rich tasks still require timely action refinement using force/torque feedback. 
Existing force-aware VLA models usually align all modalities to a single low operating frequency, which discards high-frequency contact cues that are important for reactive action correction. Additionally, they use static fusion, which cannot adaptively balance visual context and force feedback across manipulation processes. To mitigate these issues, we propose FAVLA, a force-adaptive multi-rate VLA model that explicitly separates low-frequency semantic reasoning from high-frequency force-conditioned action refinement.
Specifically, FAVLA uses a low-rate VLM to encode visual-language-force context and predict near-future force statistics, while a high-rate action expert refines action chunks using the latest force observations.
To decide \emph{when} to update actions, we introduce a Force-Adaptive Multi-rate Inference (FAMI) mechanism, which schedules the action expert's inference rate from predicted future force variance, keeping low-rate reasoning in stable phases and increasing update frequency near contact transitions.
To decide \emph{how} to use force, we design a Force-Guided Dynamic Fusion (FGDF) module, which injects high-frequency force features into the action expert and dynamically balances visual-semantic and force cues across manipulation stages. 
Extensive real-robot experiments on high-precision and contact-rich tasks show that FAVLA outperforms force-aware VLA baselines, achieving an average success rate of 88.8\% and lower peak contact forces. Zero-shot evaluations further demonstrate FAVLA’s robustness to visual distribution shifts.
Our project homepage is available at this \href{https://nmll.github.io/favla_project/}{website}.
\end{abstract}

%% file: sections/introduction.tex
\section{Introduction}

\begin{figure}[t]
    \centering
    \includegraphics[width=\columnwidth,clip,trim=10 10 10 10]{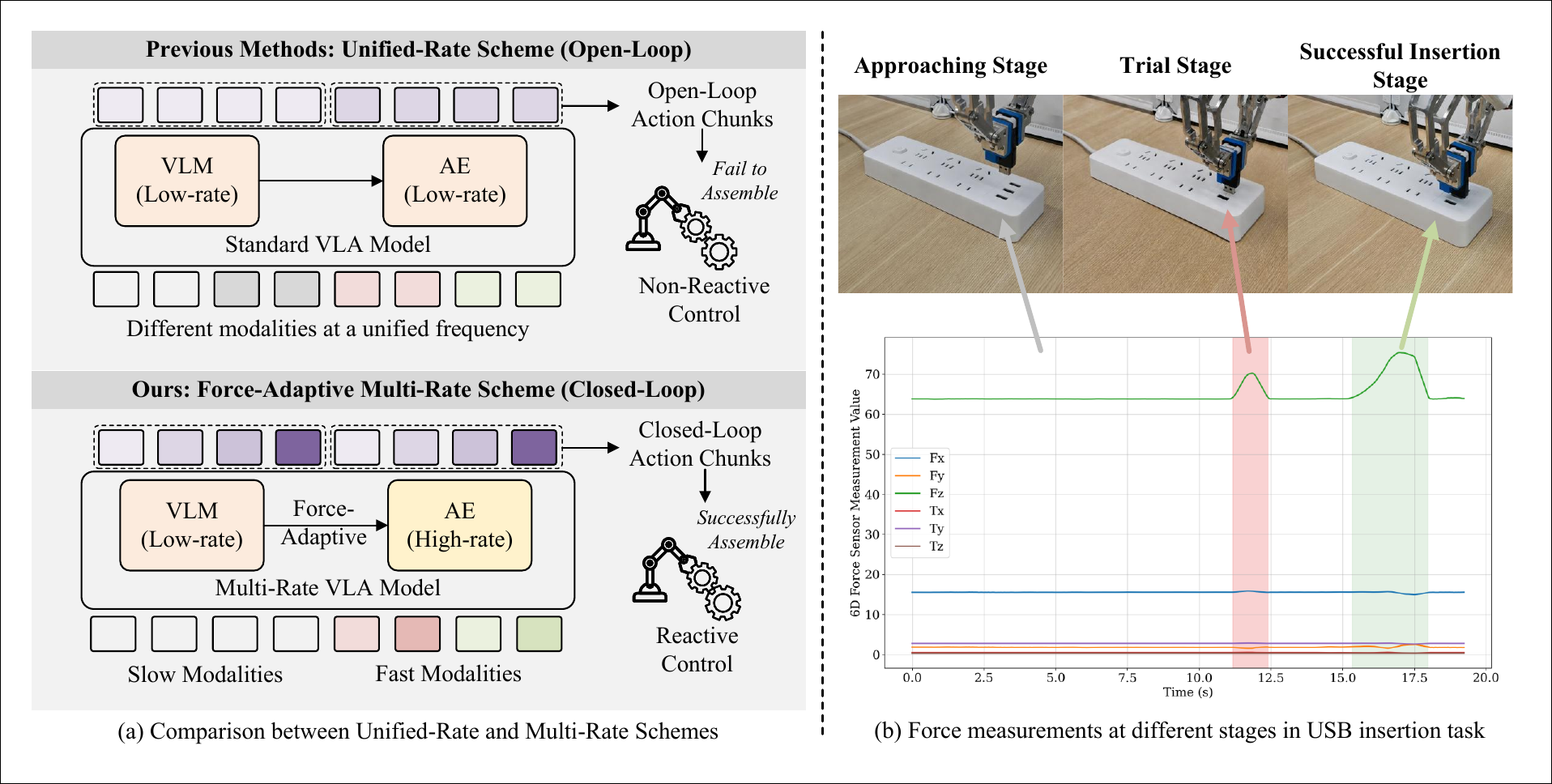}
    \caption{ (a) Existing force-aware VLA models often align all modalities to the same low frequency and generate open-loop action chunks, missing high-frequency contact cues. FAVLA instead uses a force-adaptive multi-rate scheme, allowing the action expert to refine actions with the latest force feedback. (b) Taking the USB insertion task as an example, force data exhibit different patterns across the approach, contact trial, and successful insertion stages, indicating that force sequences provide useful cues for scheduling action updates and adjusting visual-force fusion.}
    \label{fig:intro}
\vspace{-20pt}
\end{figure}

Vision-Language-Action (VLA) models have recently shown strong potential for general-purpose robotic manipulation by combining the semantic generalization of large vision-language models (VLMs) with continuous action control policies, such as $\pi_0$-style architectures~\citep{pi0,pi0.5,fast}.
However, contact-rich manipulation tasks, such as USB insertion or gear assembly, remain harder than free-space pick-and-place for general VLA models.
Success in these tasks requires both accurate visual localization and rapid action refinement under changing contact conditions.
Such conditions include abrupt impacts, stick-slip, and jamming events that are weakly observable from vision alone.

Recent work has proposed augmenting VLA or policy models with force/tactile feedback for contact-rich manipulation tasks.
For instance, ForceVLA~\citep{forcevla} uses a mixture-of-experts design to process vision and force modalities in VLA models. TA-VLA~\citep{tavla} introduces torque-aware prediction objectives as an additional constraint for action generation. Diffusion-based policies~\citep{rdp,implicitrdp} also exploit tactile or force feedback for reactive action correction.
These studies achieve effective improvements over vision-only models, demonstrating the potential of force-aware VLA models for contact-rich manipulation tasks.

Though these methods have improved manipulation performance for contact-rich tasks, how to fully exploit force feedback in VLA models remains underexplored.
There remain two critical challenges regarding \textit{when} and \textit{how} to fuse high-frequency force cues in VLA models.
First, sensors operate at different sampling frequencies (\textit{e.g.}, 15 Hz cameras \textit{vs.} 100-1000 Hz force/torque
sensors). Existing models often apply the same frequency scheme for different input modalities. They downsample high-frequency force data to match the slowest modality, which may discard critical high-frequency contact cues in force signals. 
As a result, robots fail to use these cues for timely action refinement, producing open-loop actions with excessive contact forces and  causing component damage or task failure. 
Second, token-based multimodal static fusion often suffers from imbalanced modality representation. 
Visual tokens are usually much more numerous than force tokens, so the model may overlook sparse force cues. Conversely, overemphasizing force feedback before contact can disturb visually guided target localization. 
Therefore, static fusion struggles to achieve high performance in both high-precision and stable contact-rich manipulation tasks.

To address these challenges, we propose FAVLA, a force-adaptive multi-rate VLA model based on the $\pi_0$-style architecture. By dynamically fusing high-frequency force feedback for rapid closed-loop action refinement, FAVLA achieves excellent performance on both high-precision and contact-rich manipulation tasks. 
Meanwhile, FAVLA introduces two key designs.
First, to decide \emph{when} the action expert should update actions at a higher rate, we propose a Force-Adaptive Multi-Rate Inference (FAMI) mechanism, which schedules the action expert's inference rate according to the future contact signal, enabling timely action refinement with high-frequency force data near contact transitions, while preserving low-rate VLA reasoning in stable phases. 
Meanwhile, during action refinement, we further combine fixed action sampling noise with temporal ensemble over force-refined action chunks to reduce instability caused by dynamic rate adjustment.
Second, to decide \emph{how} the action expert uses force feedback during each update, we introduce a Force-Guided Dynamic Fusion (FGDF) module to dynamically adjust visual-force fusion weights across different manipulation stages.
As shown in Fig.~\ref{fig:intro}(b), force-sequence variation implicitly reflects different manipulation stages. 
We thus use the force sequence to estimate visual-force fusion weights, without requiring additional stage annotations. 
In this way, FAVLA can emphasize visual features for accurate approach and localization, while increasing force reliance during contact and force-regulation phases for stable manipulation.

We build a real-world teleoperated robotic platform equipped with force sensors and collect a diverse manipulation dataset covering both high-precision assembly and contact-rich interaction tasks. These tasks require millimeter-level alignment, contact-state awareness, and timely force-based action refinement, such as avoiding jamming during USB insertion and gear assembly, maintaining stable contact during board wiping, box flipping, and cucumber peeling. We evaluate FAVLA against multiple baselines on the real-robot platform. FAVLA achieves an average success rate of 88.8\%, outperforming all baselines, while also reducing contact forces. Zero-shot evaluations further demonstrate FAVLA's robustness to visual distribution shifts.

In summary, this paper makes the following contributions:

 \begin{itemize}
     \item We propose FAVLA, a force-adaptive multi-rate VLA model that combines low-rate visual-language reasoning with high-rate action expert updates using force feedback for closed-loop contact-rich manipulation.
     \item We introduce a Force-Adaptive Multi-Rate Inference (FAMI) mechanism to decide \emph{when} the action expert should update at a higher rate. FAMI uses predicted future force variation to adaptively schedule the update rate across contact-transition and stable phases.
     \item We design a Force-Guided Dynamic Fusion (FGDF) module to decide \emph{how} force feedback should be used in the action expert, dynamically balancing visual-semantic cues and force cues across manipulation stages.
     \item We evaluate FAVLA on real-world high-precision and contact-rich manipulation tasks, where it outperforms force-aware VLA baselines in success rate and reduces peak contact forces.
 \end{itemize}

%% file: sections/related_work.tex
\section{Related works}

\textbf{Vision-language-action models.} Nowadays, VLA models have emerged as a powerful paradigm for learning generalist robotic policies by large-scale multimodal pretraining \citep{zhang2025upvla}. Autoregressive approaches \citep{rt-1,rt-2,openvla,fast} formulate control as next-token prediction over discretized actions, enabling scalable learning from large, heterogeneous datasets. Different from this paradigm, generative VLA models \citep{pi0,pi0.5,walloss} address action discretization by modeling continuous action trajectories via flow matching or diffusion-style objectives. More recently, hierarchical VLA architectures \citep{fast-in-slow,hirt,wen2025diffusionvla} introduce paradigms that decouple high-level reasoning from low-level execution, improving the stability and efficiency for motion planning.

\textbf{Contact-rich manipulation models.} Recent work on contact-rich manipulation extends learning-based policies by explicitly incorporating physical interaction signals beyond vision~\citep{feng2024play,liu2025forcemimic,he2025foar}. Force-aware policies \citep{rdp,implicitrdp} introduce slow–fast diffusion architectures that leverage force or force-related feedback to enable rapid corrective behaviors during contact while maintaining stable long-horizon execution. Subsequent works \citep{forcevla,tavla,cheng2025omnivtla} integrate force or torque signals into vision–language–action frameworks, allowing pretrained VLAs to adapt execution strategies based on contact states. Another line of research prefers tactile-based approaches \citep{3dvitac,liadaptivevisuotactilefusion2025,tactilevla,vlatouch} that incorporate visuo-tactile sensing to provide fine-grained contact perception and tactile-aware execution. However, existing methods typically assume the same operating frequency, which discards high-frequency cues in force data and limits reactive contact control. 
Additionally, they incorporate force data through force-aware expert routing, torque conditioning in action decoder, or predictive force-guided visuo-tactile fusion. 
FAVLA focuses on layer-wise force fusion within the action expert,
dynamically adjusting the contribution of the latest force features
across layers and manipulation stages.

In summary, 
current force-based VLA models often downsample high-frequency force signals to the same low input frequency, which may discard critical contact cues and delay action correction. 
FAVLA instead adopts a force-adaptive multi-rate strategy, using predicted force variation and latest force feedback for reactive action refinement and dynamic force fusion in contact-rich manipulation.

%% file: sections/methods.tex
\section{Methods}
\label{sec:methods}
We propose FAVLA, a force-adaptive VLA model for contact-rich manipulation, which combines force-adaptive multi-rate inference with force-guided visual-force dynamic fusion. The VLM provides low-rate visual-language context and predicts future force statistics, while the AE adaptively fuses high-frequency force feedback for reactive action refinement.
Sec.~\ref{sec:pre} formalizes contact-rich manipulation tasks with multi-frequency inputs.
Sec.~\ref{sec:multi_rate} describes the FAVLA architecture with a future force prediction head. Sec.~\ref{sec:forceadaptive} presents the Force-Adaptive Multi-Rate Inference (FAMI) mechanism. Sec.~\ref{sec:fgdf} presents the Force-Guided Dynamic Fusion (FGDF) module to dynamically balance visual-semantic and force features.

\begin{figure*}[t]
    \centering    \includegraphics[width=1.0\textwidth,clip,trim=10 10 10 10]{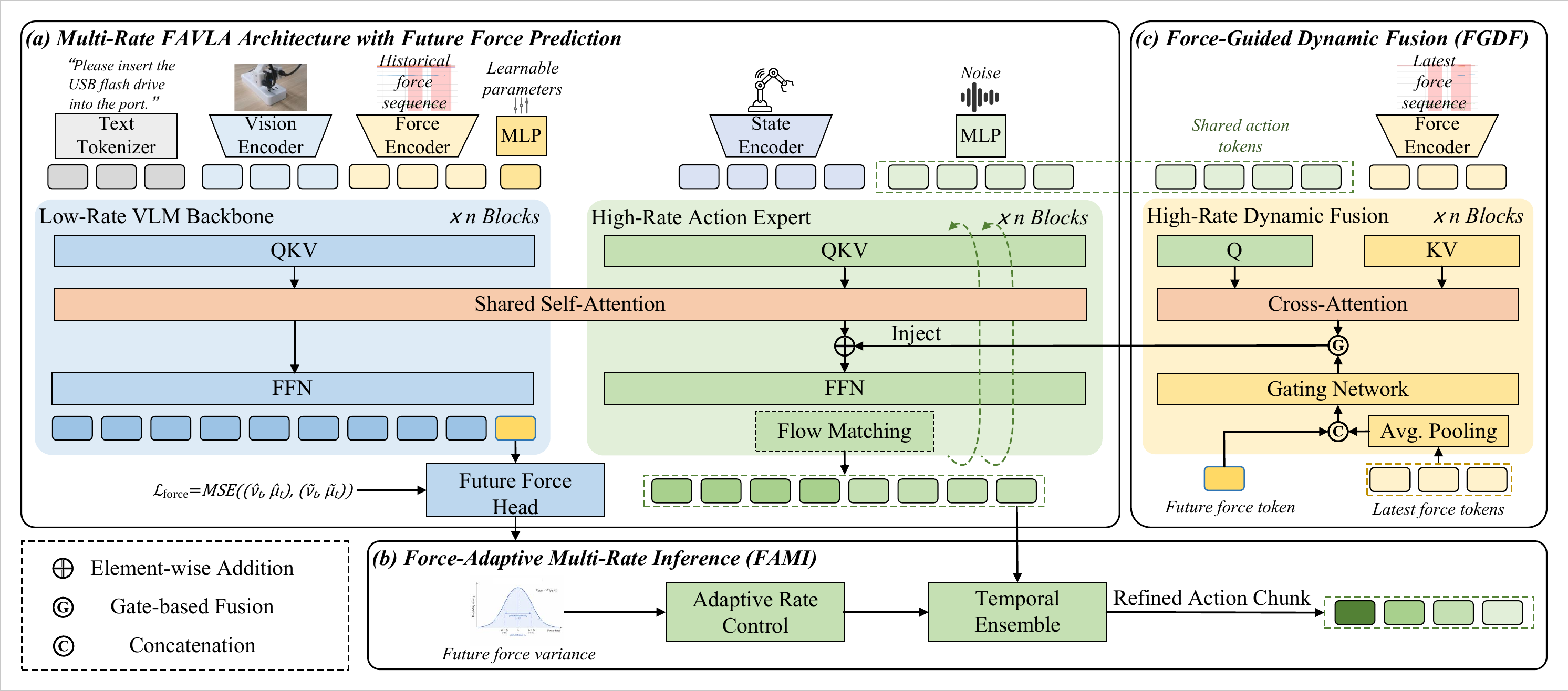}
    \caption{Overview of FAVLA. FAVLA runs the VLM and action expert at adaptive rates. The VLM encodes low-frequency vision, language, and force-history inputs and predicts future force with Force-Adaptive Multi-Rate Inference (FAMI). The action expert fuses the latest high-frequency force tokens through Force-Guided Dynamic Fusion (FGDF) and generates force-refined action chunks.}
    \label{fig:overview}
\end{figure*}

\subsection{Problem formulation}
\label{sec:pre}
We aim to build a force-adaptive multi-rate VLA model for contact-rich or high-precision industrial manipulation tasks.
Such tasks require maintaining low contact forces to avoid damaging components, which demands reactive action refinement when possible impacts occur.
To achieve this, the VLA model first needs to predict future contact force variations for early detection of impact, and then perform timely closed-loop action refinement according to high-frequency force feedback. Meanwhile, we also need to reduce action instability caused by dynamic action refinement and balance visual and force effects across manipulation stages during action generation.

Specifically, at the timestep $t$, we have the robot's observations as follows:
\begin{equation}
\label{eq:obs}
 O_t = \{\mathcal{I}_t^{(k)}\} \cup \{\mathbf{s}_t\} \cup \{\mathbf{f}_{t-\tau+1:t}, \mathbf{f}_{t'-\tau+1:t'}\},
\end{equation}
where $\mathcal{I}_t^{(k)}$ denotes the $k$-th RGB image (an external camera and a wrist camera in our setting), 
$\mathbf{s}_t\in\mathbb{R}^{7}$ is the proprioceptive state of the 6-DoF TCP (Tool Center Point of the end effector) pose and gripper width, $\mathbf{f}_{t-\tau+1:t}\in\mathbb{R}^{\tau\times 6}$ denotes a history force data sequence of 6-axis force measurements over a time window of length $\tau$, and $\mathbf{f}_{t'-\tau+1:t'}\in\mathbb{R}^{\tau\times 6}$ is the latest force data sequence at timestamp $t'$ with the same time window length $\tau$. 
In our setting, we use a 6-axis force sensor mounted on the end effector. Meanwhile, $\mathbf{f}_t=\{f_x^t, f_y^t, f_z^t, m_x^t, m_y^t, m_z^t\}$ includes the end-effector 3D force and torque components. 
In the observation, $\mathbf{s}_t$ and $\mathbf{f}_{t'-\tau+1:t'}$ are updated at a higher frequency, $\mathcal{I}_t^{(k)}$ and $\mathbf{f}_{t-\tau+1:t}$ are updated at a low frequency. 
Given a language manipulation instruction \(L\) and observation \(O_t\), we aim to learn an end-to-end VLA model \(\pi(A_t|O_t,L)\) with an auxiliary prediction head \((\nu_t,\mu_t)=h_\theta(O_t,L)\). The model predicts future force variance \(\nu_t\) and mean \(\mu_t\), together with a low-level action chunk \(A_t=\{a_t,a_{t+1},...,a_{t+H-1}\}\). The force variance captures potential future contact variations, while the force mean characterizes the expected contact force level. We use the predicted variance to represent potential future contact variations and adaptively schedule action updates. We also refine \(A_t\) with high-frequency force feedback to perform closed-loop action refinement. Each action element consists of the delta values of the 6-DoF end-effector pose and gripper state, i.e., \(a_t=(\Delta x,\Delta y,\Delta z,\Delta R_x,\Delta R_y,\Delta R_z,g)\).

\subsection{FAVLA architecture with future force prediction}
\label{sec:multi_rate}
FAVLA is designed to fuse multimodal sensory data at different sampling frequencies and predict future force statistics for adaptive inference and fusion.

\textbf{FAVLA architecture.}
Following the $\pi_0$-style VLM-AE paradigm, FAVLA uses a VLM backbone to process visual-language context and historical force information, and an action expert (AE) to generate action chunks via flow matching.
Meanwhile, RGB images and language instructions are encoded as visual and language tokens, while both the historical force sequence $\mathbf{f}_{t-\tau+1:t}$ and the latest force sequence $\mathbf{f}_{t'-\tau+1:t'}$ are encoded by a lightweight Temporal Convolutional Network (TCN) tokenizer, producing historical force tokens $\mathbf{z}_f$ and latest force tokens $\mathbf{z}'_f$, respectively.
Then the VLM concatenates low-frequency visual, language, and historical force tokens as prefix tokens. The VLM outputs Vision-Language-Force (VLF) tokens $\mathbf{H}_{\text{VLF}}$, and caches its intermediate key/value states for the AE.
The AE is conditioned on the VLM cache, robot state, and the latest high-frequency force tokens, which are further fused by the FGDF module in Sec.~\ref{sec:fgdf}.

\textbf{Future force prediction head.} 
Based on the above VLA model architecture, we add an auxiliary future force prediction head to estimate both the future force variance and mean during manipulation. 
Specifically, inspired by VLAadapter~\citep{wang2026vlaadapter}, we first append an additional learnable token to the VLM input tokens, allowing it to aggregate multimodal context from vision, language, and force-history tokens. 
Then we select the VLF token from $\mathbf{H}_{\text{VLF}}$ corresponding to the learnable input token and feed it to a MLP head to predict normalized force variance and mean values $(\hat{\nu}_t,\hat{\mu}_t)$.
The variance target is a smoothed and weighted variance of the 6D force/torque signal over a short future window defined by the action chunk size $W$.
Given the future force sequence $f_{t:t+W-1} \in \mathbb{R}^{W \times 6}$ from the dataset, we compute the raw volatility value as a weighted sum of variances:
\begin{equation}
\label{eq:raw_var}
\nu_t \;=\; \sum_{j=1}^{6} w_j\,\mathrm{Var}\big(\mathbf{f}_{t:t+W-1}^{(j)}\big).
\end{equation}
We then smooth this raw variance label with an exponential moving average (EMA) to reduce the influence of noise with
$
\bar{\nu}_t = \alpha\nu_t + (1-\alpha)\bar{\nu}_{t-1}.
$
After that, we perform the normalization operation on the $\bar{\nu}_t$ with the $tanh$ function of   
$\tilde{\nu}_t=\tanh\!\Big(\tfrac{\sqrt{{\bar{\nu}}_t}}{\sigma}\Big)$. We also normalize the future force mean target as $\tilde{\mu}_t=\mu_t/F_{\max}$, where $\mu_t$ denotes the future force mean within an action chunk, and $F_{\max}$ is the maximum force measurement range. The supervision labels are generated during dataset preprocessing.

\textbf{Training objectives.} 
We jointly optimize the VLM, the AE, and the future force predictor using a weighted sum of the action generation loss and the auxiliary force-statistics prediction loss:
\begin{equation}
\label{eq:total_loss}
\mathcal{L}_{\text{total}} = \mathcal{L}_{\text{action}} + \lambda \mathcal{L}_{\text{force}},
\end{equation}
where $\mathcal{L}_{\text{action}}$ is the flow-matching action loss following $\pi_0$, $\mathcal{L}_{\text{force}}=\|(\hat{\nu}_t,\hat{\mu}_t)-(\tilde{\nu}_t,\tilde{\mu}_t)\|^2$ is the MSE loss for predicting the normalized future force variance and mean, and $\lambda$ balances the auxiliary objective. We set $\lambda = 0.1$ in all experiments.

\subsection{Force-Adaptive Multi-Rate Inference (FAMI) mechanism}
\label{sec:forceadaptive}

We propose a Force-Adaptive Multi-rate Inference mechanism to enable high-frequency closed-loop control within each action chunk. FAMI decouples the inference rates of the VLM and AE, and adjusts the AE rate according to the predicted force variance from Sec.~\ref{sec:multi_rate}, enabling timely action refinement with high-frequency force data.

In each cycle, the VLM performs one forward pass to process multimodal observations and construct the KV cache as semantic context. Conditioned on the KV cache and the latest force observations, the AE can run multiple times for high-frequency action refinement. However, fixed high-rate AE inference cannot adapt to different task phases, and multiple adaptive AE inferences within each chunk may produce inconsistent action trajectories. We address these issues using adaptive rate scheduling and consistent temporal ensemble.

\begin{wrapfigure}[23]{r}{0.48\textwidth}
    \centering
    \includegraphics[width=\linewidth,clip,trim=10 10 10 10]{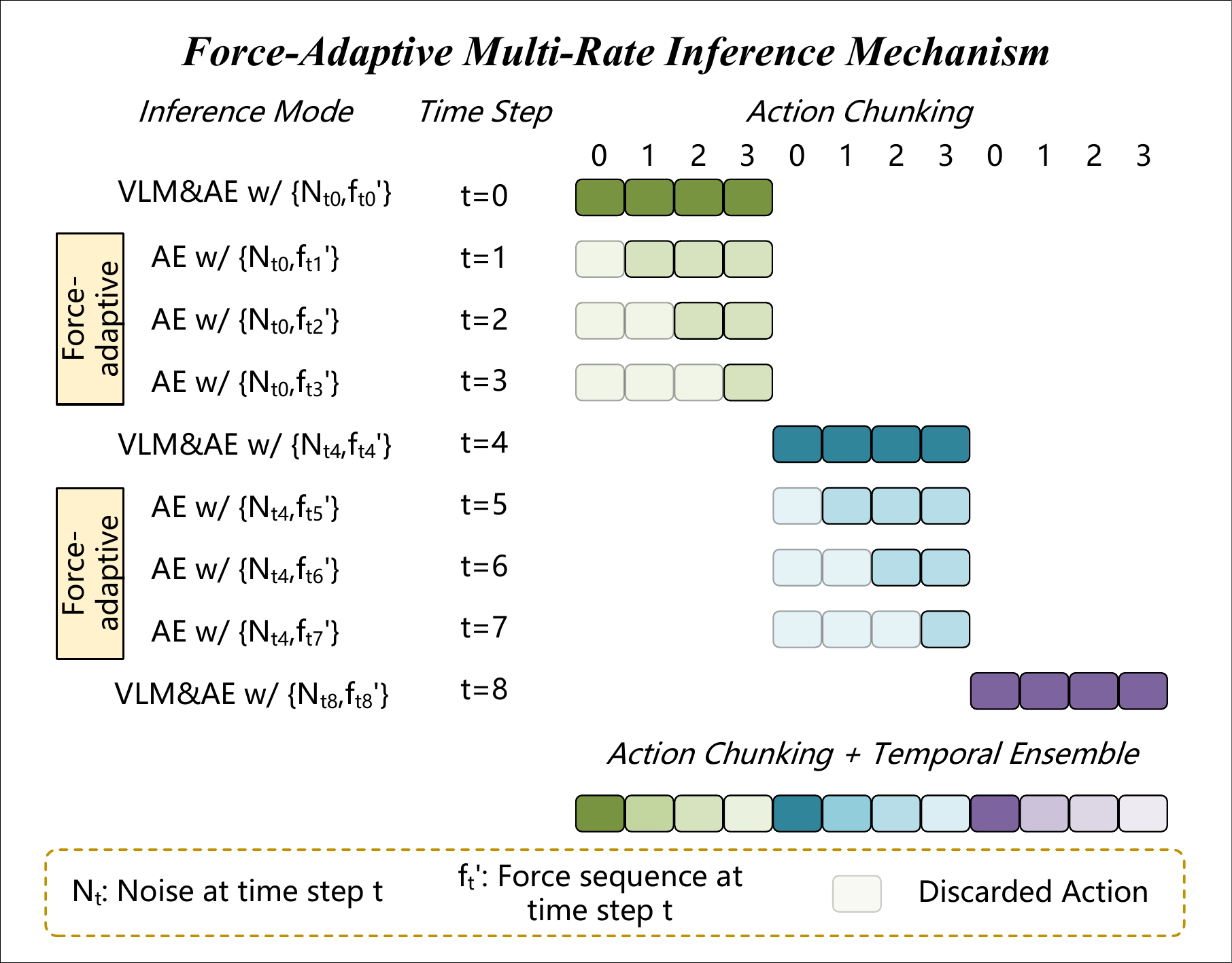}
    \caption{Our Force-Adaptive Multi-Rate Inference (FAMI) mechanism. The AE can run multiple times within each action chunk, conditioned on real-time force. In each cycle, we fix the sampled noise and robot state, then perform temporal ensemble on overlapped action chunks. The AE execution frequency is adaptively set from the VLM-predicted force variance.}
    \label{fig:force-adaptive-infer}
\end{wrapfigure}
\noindent\textbf{Force-adaptive inference mechanism.}
Rather than using a fixed high-rate inference mechanism, 
we dynamically adjust the inference rate of AE according to the future contact state. When the contact state is unstable, suggesting an impending contact event or elevated collision risk, we run AE at a high rate for timely reactive control. When the contact state is stable, \textit{i.e.}, during free-space motion or reliable  sustained contact, we run it at a low rate to ensure long-horizon action planning.

Specifically, we use the VLM's predicted future force variance $\tilde{\nu}_t \in [0,1]$ to estimate the contact state in the next cycle. Given a maximum inference rate ratio $N_{\text{max}}$, we schedule the number of AE executions as:
\begin{equation}
    n_t = \max\left(1, \left\lceil \tilde{\nu}_t \cdot N_{\text{max}} \right\rceil \right).
\end{equation}
When $\tilde{\nu}_t$ is small, we run AE at a low rate ($n_t = 1$). As $\tilde{\nu}_t$ approaches 1, more AE inferences are scheduled, providing faster response to avoid unexpected manipulation motion.

\noindent\textbf{Consistent temporal ensemble.}
To mitigate the issue of inconsistent action trajectories caused by multiple AE inferences, we adopt two strategies. First, we fix the sampled action noise $\boldsymbol{\epsilon}$ across all AE calls within the same action chunk, ensuring that all inferences share the same initial stochastic sample. Second, as shown in Fig.~\ref{fig:force-adaptive-infer}, we apply a temporal ensemble~\citep{act} to aggregate overlapped action chunks, producing smoother action execution trajectories.

\subsection{Force-Guided Dynamic Fusion (FGDF) module}
\label{sec:fgdf}
The FAMI mechanism determines \emph{when} the AE should update actions with high-frequency force feedback. 
The remaining question is \emph{how} the AE should fuse force information at each update. 
Directly injecting force with a fixed strength is suboptimal because force cues are weak or noisy before contact, but become essential during contact transitions or stable-contact phases. 
We therefore introduce Force-Guided Dynamic Fusion (FGDF), which injects the latest force features into each AE layer and dynamically balances visual-semantic features with force-sequence data. The force-sequence data implicitly reflects different manipulation stages.

Concretely, in each AE attention layer $\ell$, we first project the latest force tokens $\mathbf{z}'_f$ into force keys and values $(\mathbf{K}^{(\ell)}_f,\mathbf{V}^{(\ell)}_f)$. 
The noisy action tokens then act as queries to retrieve action-relevant force information via cross-attention:
\begin{equation}
\mathbf{A}^{(\ell)}_f
= \mathrm{Attn}\!\left(\mathbf{Q}^{(\ell)}_a,\mathbf{K}^{(\ell)}_f,\mathbf{V}^{(\ell)}_f\right),
\end{equation}
where $\mathbf{Q}^{(\ell)}_a$ is obtained from the current action-token representation. 
This action-guided attention selects force features that are most relevant to the current action refinement.

To decide how strongly force should affect the AE, FGDF builds a gate guidance signal by concatenating predicted future-force features and current-force features at the statistical level. 
Specifically, we use the VLM output token $\mathbf{h}^{\mathrm{VLM}}_t$ and pooled statistical feature of the latest force tokens:
\begin{equation}
\mathbf{g}_t =
\left[
\mathrm{sg}(\mathbf{h}^{\mathrm{VLM}}_t),
\mathrm{sg}\!\left(\mathrm{Pool}(\mathbf{z}'_f)\right)
\right],
\end{equation}
where $\mathrm{sg}(\cdot)$ denotes stop-gradient, and $\mathrm{Pool}(\cdot)$ extracts the mean and variance along the token dimension of the latest force tokens.
This provides a consistent force representation aligned with $\mathbf{h}^{\mathrm{VLM}}_t$, which is used to estimate future force mean and variance in VLM.
Each AE layer $\ell$ has an independent lightweight MLP, with parameters not shared across layers, that maps the shared guidance signal $\mathbf{g}_t$ to a scalar fusion gate through a sigmoid operation $\sigma$:
\begin{equation}
\alpha_t^{(\ell)} = \sigma\!\left(\mathrm{MLP}^{(\ell)}(\mathbf{g}_t)\right).
\end{equation}
Finally, FGDF uses $\alpha_t^{(\ell)}$ to fuse the original action representation $\mathbf{h}^{(\ell)}_a$ with the force-conditioned feature $\mathbf{A}^{(\ell)}_f$:
\begin{equation}
\mathbf{h}^{(\ell)\prime}_a
= (1-\alpha_t^{(\ell)})\mathbf{h}^{(\ell)}_a
+ \alpha_t^{(\ell)} \mathbf{A}^{(\ell)}_f .
\end{equation}
In this way, the AE can rely more on visual-semantic context during the free-space phase and more on the latest force features during contact transitions and force-dominant interactions.

%% file: sections/Experiments.tex
\section{Experiments}

\begin{figure*}[hbt]
  \begin{center}
    \centerline{\includegraphics[width=\textwidth,clip,trim=12 2 2 2]{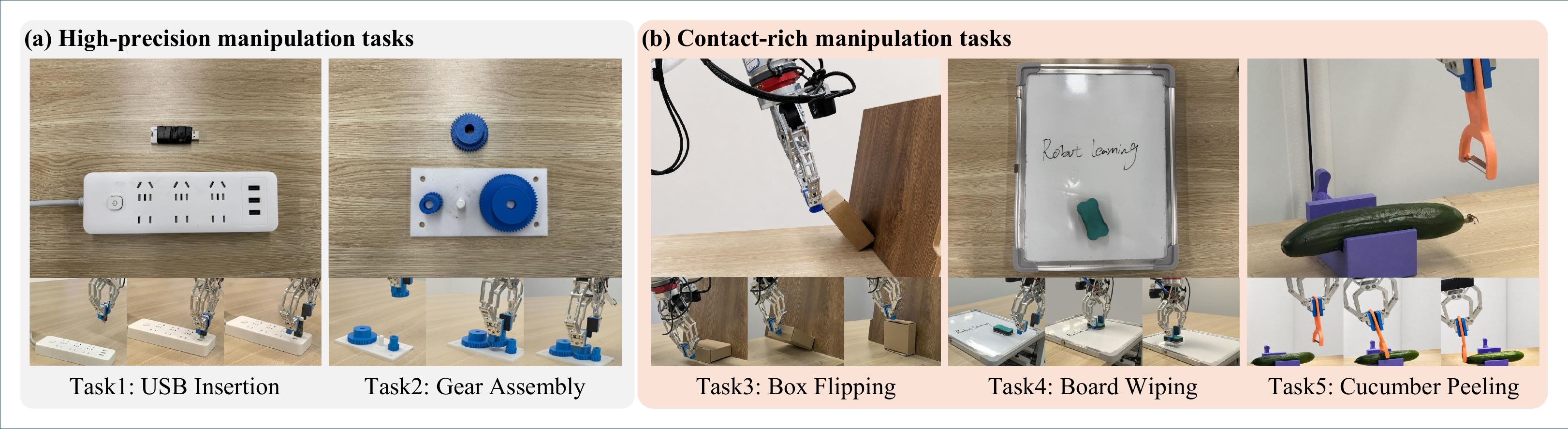}}
   \caption{Real-world manipulation tasks used in our evaluation. The tasks include high-precision manipulation tasks (USB Insertion and Gear Assembly) and contact-rich manipulation tasks (Box Flipping, Board Wiping, and Cucumber Peeling) with critical execution phases shown for each task.}
    \label{fig:taskoverview}
  \end{center}
\end{figure*}

In this section, we present a comprehensive evaluation of our proposed method on diverse real-world contact-rich tasks. We evaluate three key aspects: (1) the effectiveness of our method compared to strong baselines, (2) ablation studies validating each module's contribution, and (3) visualization results and analysis.
\subsection{Experimental setups}

\textbf{Hardware platform.}
We use a dual-arm whole-body robot with 7-DoF xARM robotic arms. Each arm is equipped with a wrist-mounted RGB-D camera and a 6-axis force/torque sensor at the end-effector flange. We capture visual observations using a wrist camera and an external fixed camera as shown in Fig.~\ref{fig:robotoverview} in the Appendix.

\textbf{Experimental tasks.} 
We evaluate our method on five real-world manipulation tasks: USB Insertion, Gear Assembly, Box Flipping, Board Wiping, and Cucumber Peeling, as shown in Fig.~\ref{fig:taskoverview}. These tasks involve different contact challenges. The first two are high-precision assembly tasks that require millimeter-level alignment and timely action refinement with force feedback to avoid jamming or excessive contact force. The last three are contact-rich manipulation tasks that require maintaining or changing contact under dynamic robot-environment interactions. 
We collect 310 human demonstration trajectories via 3D SpaceMouse teleoperation, including 80 trajectories for each high-precision task and 50 trajectories for each contact-rich task.

\textbf{Compared baselines.} 
We compare our method with the following typical baselines: $\pi_0$~\citep{pi0}, TA-VLA*, TA-VLA~\citep{tavla}, and ForceVLA~\citep{forcevla}. $\pi_0$ is a flow-matching based VLA model pretrained on large-scale data that demonstrates strong generalization on real-world tasks. TA-VLA* denotes a TA-VLA setting that concatenates force tokens into the action expert, while TA-VLA introduces an auxiliary objective for future force prediction to enhance the model's attention to force information. ForceVLA employs a Mixture-of-Experts (MoE)-based architecture~\citep{MOE} to process force and image features with modality-specific experts. All methods are fine-tuned on our collected dataset using LoRA~\citep{LoRA} for fair comparison.

\textbf{Evaluation metrics.}
We evaluate model performance using the task Success Rate (SR). For each task, we establish binary success criteria based on task completion (\textit{e.g.}, full USB insertion, correct gear alignment), and calculate SR as the ratio of successful trials to total trials. Evaluation was conducted over 25 trials for USB Insertion, 30 trials for Gear Assembly, 20 trials for Box Flipping, 20 trials for Board Wiping, and 25 trials for Cucumber Peeling. We also use average peak contact force (N) to evaluate contact force during manipulation.

\textbf{Implementation details.} 
We follow the $\pi_0$ architecture, where the VLA consists of a VLM and an action expert. The model is initialized from $\pi_0$ weights and fine-tuned with LoRA~\citep{LoRA}. All models are trained for 30,000 iterations with batch size 8 and a learning rate of $2.5 \times 10^{-5}$ under a cosine decay schedule, using a single NVIDIA A100 80GB GPU.

\begin{figure*}[tb]
\centering
\includegraphics[width=\textwidth]{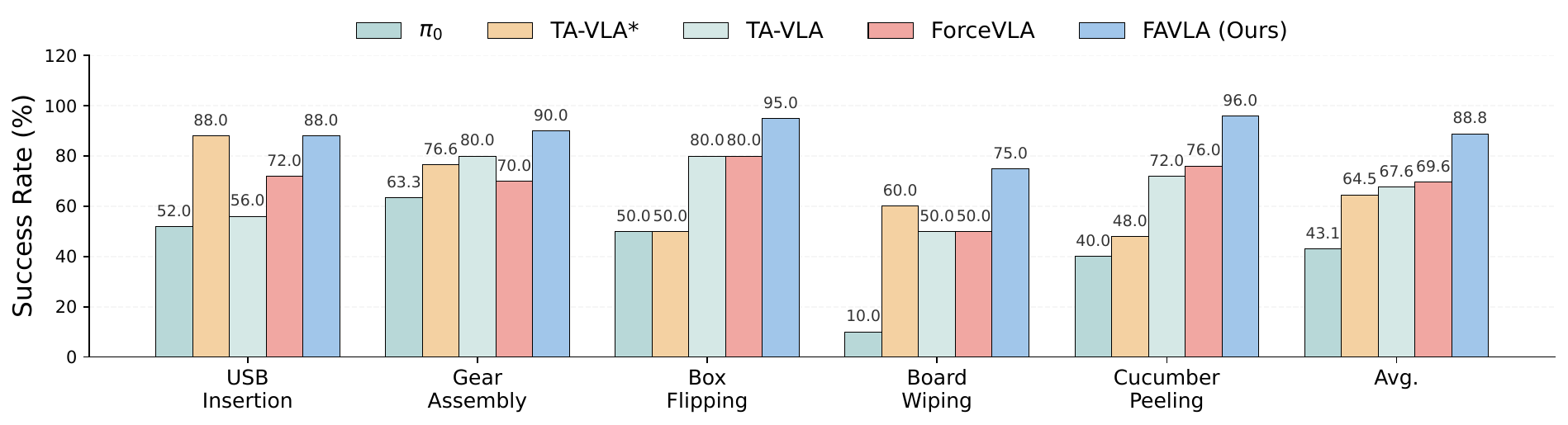}  
\caption{Comparison of the success rate of our proposed method with baselines on five real-world tasks. Our method achieves the highest average success rate and strong performance across both high-precision and contact-rich tasks.}
\label{fig:table_comparison}
\vspace{-15pt}
\end{figure*}

\subsection{Evaluation results}

\textbf{Overall performance.}
As shown in Fig.~\ref{fig:table_comparison}, FAVLA achieves the highest average performance across the five real-world manipulation tasks. Our method achieves an average success rate of 88.8\%, outperforming the vision-only $\pi_0$
 baseline by 45.7 percentage points and surpassing the best baseline ForceVLA by 19.2 percentage points. These results demonstrate the effectiveness of our approach.

We observe significant improvements on Gear Assembly (90.0\%), Box Flipping (95.0\%), Board Wiping (75.0\%), and Cucumber Peeling (96.0\%). 
These tasks are sensitive to contact dynamics: Gear Assembly requires accurate alignment and timely action refinement, as even slight misalignment can cause collisions between parts and potential damage.
Box Flipping, Board Wiping, and Cucumber Peeling require maintaining appropriate contact force throughout the motion. 
Our FAVLA addresses this by providing closed-loop action control and dynamically balancing visual-semantic and force cues, enabling reactive action adjustments based on high-frequency force feedback.

\begin{wraptable}[13]{r}{0.6\textwidth}
\centering
\vspace{-10pt}
\caption{Average peak contact forces (N) on representative tasks that require low contact forces. Lower values indicate reduced excessive contact and more compliant manipulation.}
\label{tab:force_analysis}
\renewcommand{\arraystretch}{1.3}
\setlength{\tabcolsep}{3pt}
\begin{tabular*}{\linewidth}{l @{\extracolsep{\fill}} cc}
\toprule
\textbf{Method} & \textbf{Gear Assembly} & \textbf{Box Flipping} \\
\midrule
$\pi_0$               & 12.5 N & 12.8 N \\
TA-VLA*         & 13.0  N  & 13.9  N  \\
TA-VLA                & 13.5  N  & 11.0 N  \\
ForceVLA              & 10.9  N  & 12.6  N  \\
\midrule
\textbf{FAVLA (Ours)} & \textbf{7.6 N } & \textbf{10.9 N } \\
\bottomrule
\end{tabular*}
\end{wraptable}

\textbf{Capabilities of force control.}  We further validate the force control capability by measuring average peak contact forces. This metric captures the maximum contact force at the end effector during each trial, averaged across all trials per task. We select Box Flipping and Gear Assembly tasks for evaluation because the gear and cardboard box can easily break under large manipulation forces.

As shown in Tab.~\ref{tab:force_analysis}, FAVLA achieves lower peak contact forces on the two tasks, with 7.6~N on Gear Assembly and 10.9~N on Box Flipping. Compared with the vision-only baseline, this corresponds to reductions of 4.9~N and 1.9~N, respectively. Although force-aware baselines (TA-VLA and ForceVLA) fuse  additional force data, they still produce relatively high contact forces on at least one task.
By contrast, with the force-adaptive multi-rate inference strategy, FAVLA can react to contact changes using high-frequency force feedback, reducing excessive contact during manipulation.

\textbf{Out-of-distribution (OOD) Evaluation.}
We evaluate the original task-trained models under OOD conditions without additional demonstrations or fine-tuning (Tab.~\ref{tab:ood_generalization}). Gear Assembly tests unseen gear colors and assembly-base heights. Both tasks additionally test unseen geometry, changed backgrounds, and external-camera viewpoints. For unseen geometry, we use a smaller gear with a diameter of 1.5~cm while retaining the original assembly base, and a geometrically distinct, lower-friction workpiece for Box Flipping. Background tests retain in-domain objects while changing background colors. Viewpoint tests relocate and reorient the external camera while keeping the wrist camera fixed. Task instructions and the robot platform remain unchanged. 
FAVLA achieves higher SR than TA-VLA across all eight settings and lower peak contact forces except on the lower-friction flipping workpiece (15.2 vs  14.6~N). These results suggest that FAVLA can leverage force feedback through the FGDF module to complete OOD tasks, rather than relying solely on visual information.

\begin{table}[t]
    \centering
    \caption{OOD evaluation results. SR denotes the success rate (\%) and MaxF represents the average peak contact force (N).}
    \label{tab:ood_generalization}
    \setlength{\tabcolsep}{3pt}
    \renewcommand{\arraystretch}{1.15}
    \begin{tabular*}{\linewidth}{@{\extracolsep{\fill}}llrrrr@{}}
        \toprule
        \multirow{2}{*}{\textbf{Task}} & \multirow{2}{*}{\textbf{OOD Setting}} & \multicolumn{2}{c}{\textbf{TA-VLA}} & \multicolumn{2}{c}{\textbf{FAVLA}} \\
        \cmidrule(lr){3-4}\cmidrule(l){5-6}
        & & \textbf{SR} ($\uparrow$) & \textbf{MaxF} ($\downarrow$) & \textbf{SR} ($\uparrow$) & \textbf{MaxF} ($\downarrow$) \\
        \midrule
        \multirow{5}{*}{Gear Assembly} & Unseen colors & 70.0 & 14.4 & \textbf{75.0} & \textbf{9.6} \\
        & Placement height & 60.0 & 15.9 & \textbf{80.0} & \textbf{9.5} \\
        & Unseen geometry & 30.0 & 20.6 & \textbf{60.0} & \textbf{13.6} \\
        & Background & 53.3 & 14.4 & \textbf{90.0} & \textbf{8.2} \\
        & Camera viewpoint & 53.3 & 17.8 & \textbf{70.0} & \textbf{12.2} \\
        \midrule
        \multirow{3}{*}{Box Flipping} & Unseen geometry & 40.0 & \textbf{14.6} & \textbf{50.0} & 15.2 \\
        & Background & 80.0 & 13.6 & \textbf{95.0} & \textbf{10.5} \\
        & Camera viewpoint & 65.0 & 14.4 & \textbf{80.0} & \textbf{10.1} \\
        \bottomrule
    \end{tabular*}
\end{table}

\begin{wraptable}[14]{r}{0.6\textwidth}
\centering
\vspace{-10pt}
\caption{Ablation study on proposed components. We report the success rate (SR) and peak contact force (MaxF, N) for each configuration.}
\label{tab:ablation_rigorous}
\renewcommand{\arraystretch}{1.3}
\setlength{\tabcolsep}{2pt}
\begin{tabular*}{\linewidth}{@{}l @{\extracolsep{\fill}} cc cc@{}}
\toprule
\multirow{2}{*}{\textbf{Configuration}} & \multicolumn{2}{c}{\textbf{Gear Assembly}} & \multicolumn{2}{c}{\textbf{Box Flipping}} \\
\cmidrule(lr){2-3}\cmidrule(l){4-5}
 & \textbf{SR} ($\uparrow$) & \textbf{MaxF} ($\downarrow$) & \textbf{SR} ($\uparrow$) & \textbf{MaxF} ($\downarrow$) \\
\midrule
Vision-Only & 63.3\% & 12.5 N & 50.0\% & 12.8 N \\
$w/\;$ Force Pred. & 70.0\% & 13.6 N & 70.0\% & 11.7 N \\
$w/\;$ FAMI & 73.3\% & 12.0 N & 80.0\% & \textbf{10.1 N} \\
$w/\;$ FGDF & \textbf{90.0\%} & \textbf{7.6 N} & \textbf{95.0\%} & 10.9 N \\
\bottomrule
\end{tabular*}
\end{wraptable}

\subsection{Ablation studies}
\textbf{Ablation study on proposed components.} To validate our architectural design, we conduct an ablation study by progressively integrating each proposed component into a vision-only baseline, as shown in Tab.~\ref{tab:ablation_rigorous}. 
(1) The Vision-Only baseline has limited performance on these two tasks with 63.3\% and 50.0\% SRs, respectively. 
(2) Adding future force prediction improves the SR, suggesting that future force information provides an implicit cue about upcoming action intent and thus assists action generation. However, this module alone does not reduce contact force.
(3) With FAMI, the model further improves SR, and especially reduces the peak contact forces to 12.0~N and 10.1~N on the two tasks, respectively. This shows that timely action refinement with high-frequency force data can help the model respond to contact changes and reduce excessive contact force.
(4) Finally, adding FGDF achieves the best success rates, 90.0\% on Gear Assembly and 95.0\% on Box Flipping. Meanwhile, we obtain a 7.6~N peak contact force on the gear assembly task. We observe that the model performs more accurate localization, which reduces collisions between the loose gear and the assembly base. This suggests that FGDF helps preserve visual alignment cues while incorporating force feedback for contact refinement.
On Box Flipping, FGDF also improves SR by helping the model attend to force-conditioned action features, enabling more stable box contact during the flipping motion.

\subsection{Visualization analysis of FAMI and FGDF}

\begin{figure*}[t]
\centering
\begin{minipage}{0.49\textwidth}
    \centering
    \includegraphics[width=\linewidth]{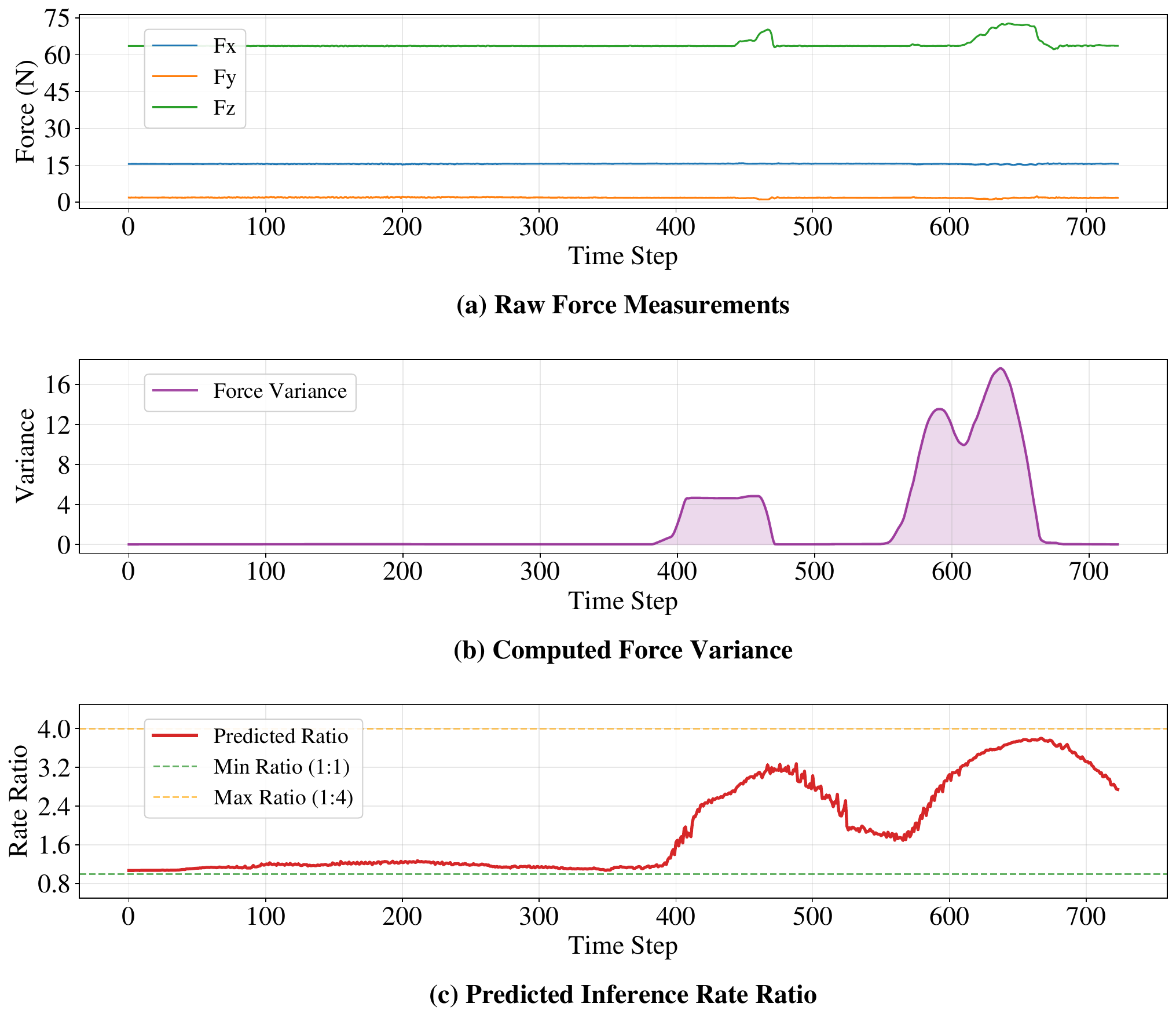}
    \caption{Visualization of predicted inference rate ratio in FAMI. (a) Raw force measurements during USB Insertion. (b) Force variance calculated from force observations to indicate time steps with large force variations. (c) Predicted inference rate ratio in FAMI, showing time steps where the AE update frequency is increased.}
    \label{fig:freq_prediction}
\end{minipage}
\hfill
\begin{minipage}{0.47\textwidth}
    \centering
    \includegraphics[width=\linewidth]{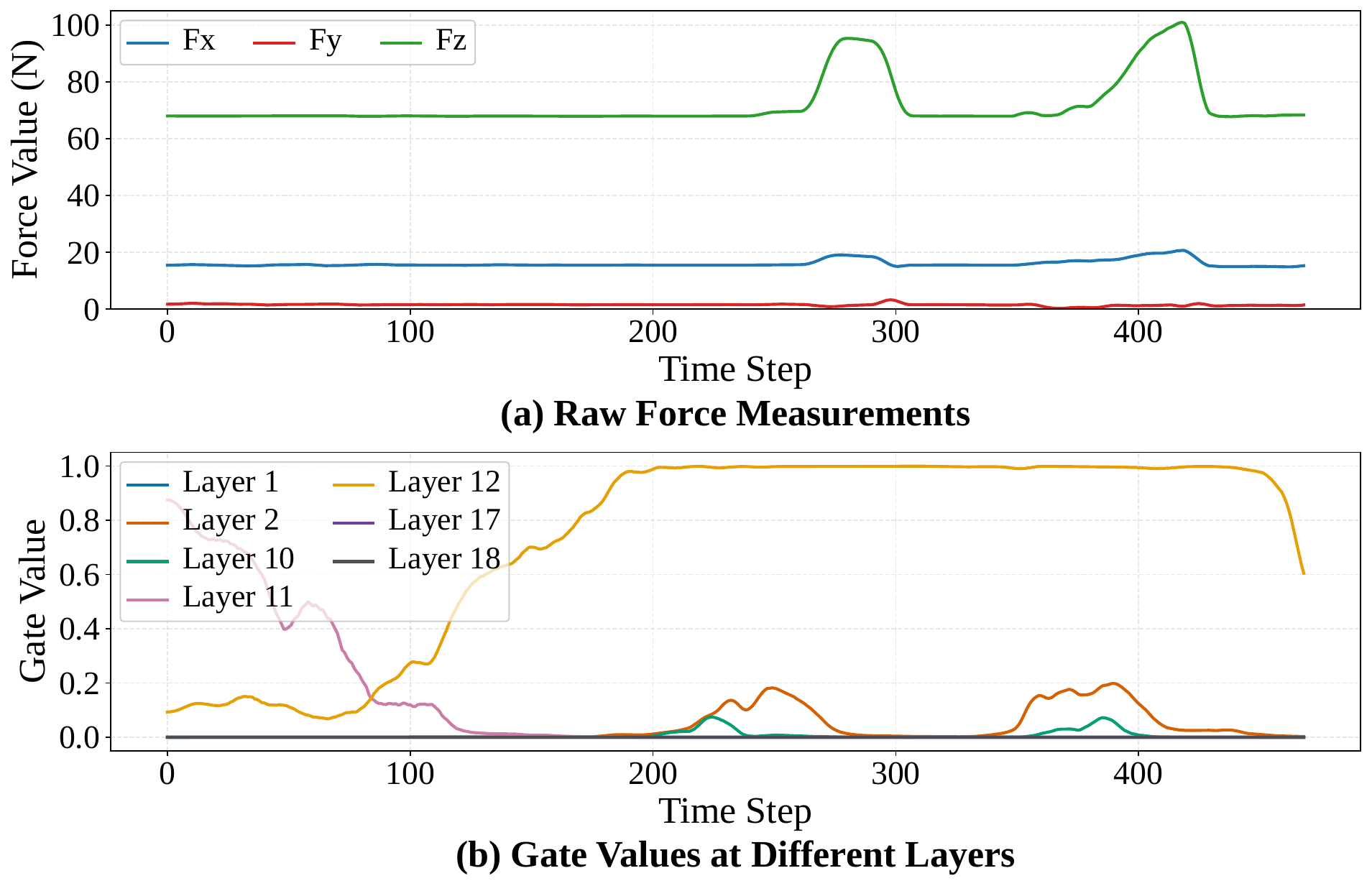}
    \caption{Visualization of force gate values at different layers predicted in FGDF. (a) Raw force measurements during USB insertion. (b) Predicted gate values of force-conditioned action features at different layers in AE. The gates show selective activation across layers and time steps: shallow layers (Layer1-2) show transient responses, middle layers (Layer10-12) are activated more persistently during force-sensitive phases, while the final layers (Layer17-18) remain mostly inactive.}
    \label{fig:fgdf_gate_visualization}
\end{minipage}
\vspace{-18pt}
\end{figure*}

Fig.~\ref{fig:freq_prediction}  visualizes FAMI's predicted inference rate during a USB insertion sequence.
Fig.~\ref{fig:freq_prediction}(a) shows the raw force measurements, they remain relatively stable in the early stage and show clear fluctuations during the trial or insertion processes. Fig.~\ref{fig:freq_prediction}(b) shows the force variance calculated from the force observations, which shows these time steps with large force variations. 
Correspondingly, in Fig.~\ref{fig:freq_prediction}(c), FAMI predicts increased inference rate ratios around these periods, while keeping a lower rate ratio during stable phases. This behavior is consistent with our expectation that the AE should update more frequently when contact dynamics become significant, while avoiding unnecessary high-frequency inference during stable interactions.
Fig.~\ref{fig:fgdf_gate_visualization} shows the visualization results of gate values on force-conditioned action features at different layers in FGDF. 
From Fig.~\ref{fig:fgdf_gate_visualization}(b), we see that shallow layers such as Layer 2 mainly exhibit local or transient responses, which may correspond to early action encoding. Large force variations often indicate the need for action adjustment, so transient force injection at shallow layers may help modulate the early action representation before it is further refined by deeper layers. 
Middle layers, especially Layers 11-12, show stronger and more sustained activation, suggesting that force cues are primarily fused in middle layers for action adjustment. 
The low gate values in the final layers suggest that force information is injected before the final output refinement stage, while the last layers tend to preserve the consolidated action representation.

%% file: sections/Conclusion.tex
\section{Conclusion}

In this paper, we present FAVLA, a force-adaptive VLA framework designed to fuse “slow” semantic modalities (vision, language, and force history) with “fast” proprioceptive state and force signals into a single end-to-end VLA model. 
Our multi-rate architecture employs a large VLM to extract vision–language–force context and predict future contact force variations, while a lightweight action expert leverages high-frequency force measurements for reactive action refinement. 
With this architecture, we introduce the Force-Adaptive Multi-rate Inference mechanism to adjust the action expert's execution frequency according to predicted force variation, enabling timely action updates near contact transitions.
We further introduce the Force-Guided Dynamic Fusion module, which dynamically balances visual-semantic and force cues in the action expert, preserving visual guidance for alignment while injecting force information for contact refinement.
Extensive real-robot experiments on five high-precision and contact-rich manipulation tasks show that FAVLA achieves an $88.8\%$ average success rate, outperforming the best baseline by $19.2$ percentage points, especially reduces peak contact forces. These results demonstrate that force-adaptive multi-rate inference and dynamic visual-force fusion are effective for high-precision and contact-rich manipulation tasks.

%% file: sections/Appendix.tex
\section{Appendix}

{\setlength{\intextsep}{5pt}
\begin{wrapfigure}[23]{r}{0.4\textwidth}
    \centering
    \vspace{-6pt}
    \includegraphics[width=0.9\linewidth,clip,trim=12 2 12 2]{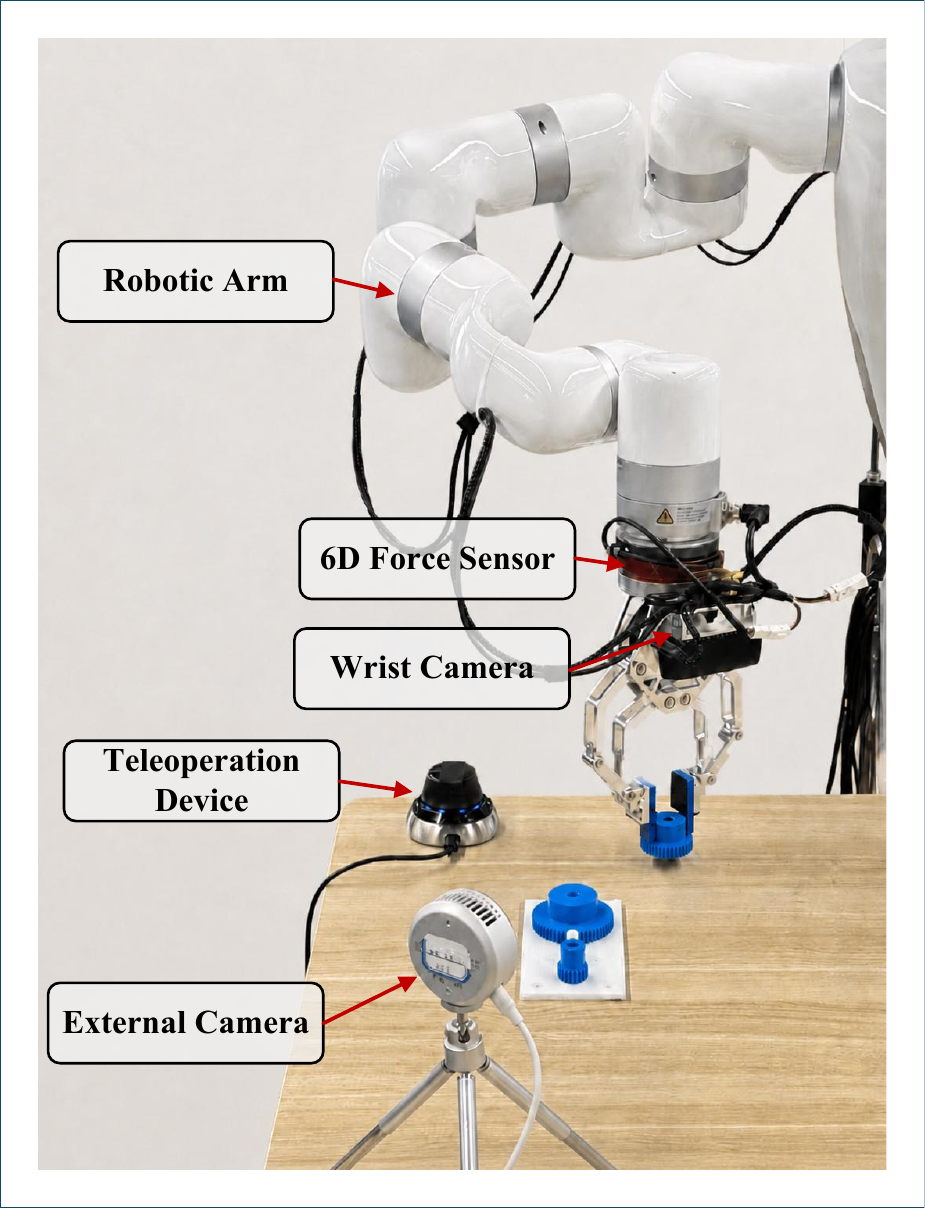}
    \caption{
    Hardware setup for robotic manipulation. The platform includes a 7-DOF X-ARM arm, a wrist-mounted camera, an external camera, a 6D force sensor at the wrist, and a 3D SpaceMouse teleoperation device for collecting demonstration data.}
    \label{fig:robotoverview}
\end{wrapfigure}

\subsection{Teleoperation system}
\label{sec:teleoperation}
Our teleoperation system is shown in Fig.~\ref{fig:robotoverview}. An operator controls the robot in end-effector space using a 3D SpaceMouse. The relative motion of the SpaceMouse is scaled and mapped to the robot’s relative end-effector motion.
To address the lack of force feedback, we convert the force readings into audio signals. Each of the 6 force dimensions is mapped to a distinct audio frequency (220-880 Hz), enabling the operator to identify which direction is experiencing force. The sound volume is proportional to the force magnitude. The operator wears headphones to perceive both the direction and magnitude of contact forces through real-time audio feedback.

\subsection{Data collection details}
We collect demonstration data through human teleoperation using the system described in Sec.~\ref{sec:teleoperation}. A human operator controls the robot arm via a 3D SpaceMouse to complete manipulation tasks, while the system simultaneously records trajectory data, including robot states, force/torque measurements, and camera observations.

To increase trajectory diversity, we randomize the initial positions of both the robot arm and task objects at the beginning of each demonstration. We find that varying positions helps the model learn more generalizable manipulation strategies.

During teleoperation, we record multiple data streams at different frequencies: end-effector poses and 6D force/torque readings at 200 Hz, along with external camera and wrist camera images at 30 Hz with a resolution of $640\times360$. All streams are synchronized and downsampled to 30 Hz for temporal alignment. The collected dataset is then converted to the LeRobot v2.1 format for subsequent training.

\subsection{Dataset statistics}
We present dataset statistics for our collected datasets in Tab.~\ref{tab:dataset_stats}, including the number of trajectories and frames per task. Our dataset comprises five real-world manipulation tasks, with a total of 310 trajectories and 228,352 frames.
We use these demonstrations for task-specific fine-tuning of pretrained VLA models, primarily to evaluate the effectiveness of FAMI and FGDF.

\vspace{8pt}
\begin{table}[h]
\centering
\caption{Dataset statistics for the five manipulation tasks.}
\label{tab:dataset_stats}
\renewcommand{\arraystretch}{1.3}
\begin{tabular}{lcc}
\toprule
\textbf{Task} & \textbf{Trajectories} & \textbf{Frames} \\
\midrule
USB Insertion & 80 & 44,716 \\
Gear Assembly & 80 & 58,038 \\
Box Flipping & 50 & 47,476 \\
Board Wiping & 50 & 48,020 \\
Cucumber Peeling & 50 & 30,102 \\
\midrule
\textbf{Total} & \textbf{310} & \textbf{228,352} \\
\bottomrule
\end{tabular}
\end{table}

\begin{table}[t]
\centering
\caption{Specification of the observation and action spaces.}
\label{tab:observation_space}
\renewcommand{\arraystretch}{1.2}
    \begin{tabular}{lcc}
    \toprule
    \textbf{Field} & \textbf{Components} & \textbf{Shape} \\
    \midrule
    \multicolumn{3}{l}{\textbf{Action (Chunk Size = 32)}} \\
    Delta End-Effector Pose + Gripper Width
    & $\Delta x, \Delta y, \Delta z, \Delta \phi, \Delta \theta, \Delta \psi, g$ 
    & $32 \times 7$ \\
    \midrule
    \multicolumn{3}{l}{\textbf{Observation}} \\
    End-Effector Pose + Gripper Width
    & $x, y, z, \phi, \theta, \psi, g$ 
    & 7 \\
    Force / Torque History (10 Steps) 
    & $f_x, f_y, f_z, \tau_x, \tau_y, \tau_z$ 
    & $10 \times 6$ \\
    Head Image
    & RGB 
    & $360 \times 640 \times 3$ \\
    Wrist Image
    & RGB 
    & $360 \times 640 \times 3$ \\
    Text Instruction
    & Tokenized Text Sequence
    & 48 \\
    \bottomrule
    \end{tabular}
\end{table}

\subsection{Task description}
\textbf{USB Insertion.}
The goal of this task is to fully insert a USB drive into a standard USB port. The robot initially holds a USB drive in its gripper. To successfully complete the task, it must first move above the target port, align the drive with the port, and finally perform the insertion. This task demands sub-millimeter precision, as even slight misalignment can result in insertion failure.

\textbf{Gear Assembly.}
The goal is to mount a gear held by the robot gripper onto a central pillar on the base, ensuring its teeth fully mesh with two pre-installed adjacent gears on both sides. This task presents two main challenges:
First, the gear’s hole must be aligned with the pillar. Since the hole is not visible, the robotic arm must rely on real-time force feedback to determine alignment.
Second, for proper meshing, the robot also needs to use force feedback to ensure the gear is aligned during the final engagement.

\textbf{Box Flipping.}
The objective of this task is to flip a horizontally placed cardboard box to a vertical orientation. This task requires the robot to adapt to constantly changing contact forces and, based on force feedback, adjust the direction and magnitude of the applied force to avoid damaging the cardboard box with excessive force or causing it to slip due to insufficient contact.

\textbf{Board Wiping.}
This task requires the robot to grasp the eraser and erase the markings on the board. The challenge lies in ensuring stable contact between the eraser and the board surface to apply sufficient force for effective erasure.

\textbf{Cucumber Peeling.}
The goal of this task is to peel the cucumber skin into continuous strips using a handheld peeler. This task requires sustained contact between the peeler and the cucumber surface while regulating the applied force: excessive force may damage the cucumber or interrupt the motion, whereas insufficient force leads to incomplete peeling.

\begin{table}[t]
    \centering
    \caption{Summary of model architecture and hyperparameters.}
    \label{tab:model_architecture}
    \renewcommand{\arraystretch}{1.2}
    \begin{tabular}{@{}p{0.28\linewidth}p{0.66\linewidth}@{}}
        \toprule
        \textbf{Module} & \textbf{Configuration} \\
        \midrule
        \multicolumn{2}{@{}l}{\cellcolor{gray!25}\textbf{Vision-Language Model}} \\
        Total Parameters            & 2.6B \\
        Transformer Layers          & 18 \\
        Hidden Dimension           & 2048 \\
        FFN Hidden Dimension       & 16384 \\
        Attention Mechanism        & Grouped Query Attention, 8 query heads, 1 key-value head, head\_dim = 256 \\
        LoRA                        & rank $r=16$, scaling factor $\alpha=16$ \\
        \midrule\addlinespace[0.4em]
        \multicolumn{2}{@{}l}{\cellcolor{gray!25}\textbf{Action Expert}} \\
        Total Parameters            & 300M \\
        Transformer Layers          & 18 \\
        Hidden Dimension           & 1024 \\
        FFN Hidden Dimension       & 4096 \\
        Attention Mechanism        & Grouped Query Attention, 8 query heads, 1 key-value head, head\_dim = 256 \\
        LoRA                        & rank $r=32$, scaling factor $\alpha=32$ \\
        \midrule\addlinespace[0.4em]
        \multicolumn{2}{@{}l}{\cellcolor{gray!25}\textbf{FGDF Model}} \\
        Fusion Layers              & All 18 action-expert layers \\
        Gate MLP Dimensions        & $4098, 1024, 1$ \\
        Gate Activation            & Swish, Sigmoid \\
        \midrule\addlinespace[0.4em]
        \multicolumn{2}{@{}l}{\cellcolor{gray!25}\textbf{TCN Force Encoder}} \\
        Total Parameters            & 25.2M \\
        TCN Blocks                 & 4 \\
        Hidden Dimension           & 1024 \\
        Convolution Kernel Size    & 3 \\
        Dilation Rates             & $[1, 2, 4, 8]$  \\
        Activation Function        & Swish \\
        Number of Output Force Tokens    & 4 \\
        \midrule\addlinespace[0.4em]
        \multicolumn{2}{@{}l}{\cellcolor{gray!25}\textbf{Force Variance Predictor}} \\
         Total Parameters            & 1.6M \\
            MLP Hidden Dimensions      & $(256, 128)$ \\
            Input Dimension            & 6144\\
            Output Dimension           & 2 \\
            Activation Function        & GELU \\
        \bottomrule
    \end{tabular}
\end{table}

\subsection{Training and implementation details}
The observation and action spaces are defined in Tab.~\ref{tab:observation_space}. The robot's input consists of multi-modal observations, including RGB images from wrist and external cameras, proprioceptive state, 6-axis force/torque data, and language instructions. The output action space consists of delta 6-DoF end-effector poses and gripper width. The model predicts an action chunk with a chunk size of 32.

We present the model architecture and parameters in Tab.~\ref{tab:model_architecture}. We adopt the $\pi_0$ architecture, which integrates a Vision-Language Model (VLM) and an action expert, and add a new force variance network to predict the force variance. 
The VLM and action expert are initialized with $\pi_0$ pre-trained weights and are fine-tuned using LoRA, while the force variance network is trained from scratch.
The LoRA parameters are configured with $r=16$ and $\alpha=16$ for the VLM, and $r=32$ and $\alpha=32$ for the action expert, applied to both attention and feedforward layers.

During inference, the action expert uses 10 denoising steps. FAVLA adaptively selects the VLM:AE execution ratio from 1:1 to 1:4 according to the predicted force variance, where the maximum ratio is used when contact variation is high. For ablation variants without FAMI, the AE runs at a fixed 1:1 frequency. We compute the force-variance target using dimension-wise weights $[1.0, 1.0, 1.0, 0.5, 0.5, 0.5]$ for the 3D force and 3D torque components, and smooth the target over a short future window to stabilize training.

We measure inference latency on an NVIDIA RTX 4090 GPU. The VLM latency is approximately 47 ms, and the AE latency is approximately 27 ms with 10 denoising steps and action decoding. Under the 1:1 setting, one VLM-AE execution cycle takes approximately 74 ms, corresponding to a 13.5 Hz action update frequency. Under the 1:4 setting, one cycle takes approximately 155 ms, corresponding to a 6.5 Hz VLM update frequency and a 25.8 Hz AE action update frequency.

We train a task-specific model on each dataset for 30,000 iterations with a batch size of 8, using the AdamW optimizer and a cosine decay learning rate schedule. Training takes approximately 8 hours on a single NVIDIA A100 80GB GPU.

\subsection{Supplementary analysis}

\subsubsection{Comparison with simpler scheduling strategies}
\label{sec:scheduling_comparison}
We additionally evaluate the contribution of FAMI by replacing its predictive scheduler with fixed VLM:AE update ratios of 1:1, 1:2, and 1:4, a force-threshold scheduler, and a current-force-variance scheduler. All variants retain FGDF, so these comparisons change scheduling rather than force fusion.
The force-threshold scheduler increases the AE update rate when the current force magnitude exceeds 5~N. The current-variance scheduler increases it when the variance in the latest observed force window exceeds 4.0. The thresholds are selected from demonstration force statistics and held fixed across tasks. The heuristic thresholds yield average AE update frequencies comparable to FAMI.

\begin{table}[htbp]
  \centering
  \caption{Comparison with simpler scheduling strategies. All variants retain FGDF.}
  \label{tab:scheduling_comparison}
  \setlength{\tabcolsep}{3pt}
  \begin{tabular*}{\linewidth}{@{}l@{\extracolsep{\fill}}rrrr@{}}
    \toprule
    & \multicolumn{2}{c}{Gear Assembly} & \multicolumn{2}{c}{Box Flipping} \\
    \cmidrule(lr){2-3}\cmidrule(l){4-5}
    Scheduler & SR $\uparrow$ & MaxF $\downarrow$ & SR $\uparrow$ & MaxF $\downarrow$ \\
    \midrule
    Fixed 1:1 & 76.6 & 9.2 & 85.0 & 14.1 \\
    Fixed 1:2 & 83.3 & 8.7 & \textbf{95.0} & 11.4 \\
    Fixed 1:4 & 86.6 & 8.5 & \textbf{95.0} & 11.9 \\
    Force threshold & 83.3 & 9.0 & \textbf{95.0} & 11.1 \\
    Current-force variance & \textbf{90.0} & 8.6 & \textbf{95.0} & 11.8 \\
    Predicted FAMI & \textbf{90.0} & \textbf{7.6} & \textbf{95.0} & \textbf{10.9} \\
    \bottomrule
  \end{tabular*}
\end{table}

Tab.~\ref{tab:scheduling_comparison} shows that FAMI matches or improves the highest SR and has the lowest peak force on both tasks. Relative to fixed 1:4, it improves Gear Assembly SR from 86.6\% to 90.0\% and reduces peak force from 8.5 to 7.6~N. It matches the 90.0\% SR of the current-variance scheduler with lower peak force (7.6 vs.\ 8.6~N). On Box Flipping, several schedulers reach 95.0\% SR, but FAMI achieves the lowest peak force of 10.9~N.
Unlike the reactive heuristics, which respond after force changes are observed, FAMI uses predicted future variation to enhance refinement before contact transitions.

\subsubsection{Comparison with fixed fusion gates}
\label{sec:fixed_fusion_gates}
We replace learned FGDF gates with constant force weights $\alpha\in\{0.2,0.5,0.8\}$ to assess whether a manually selected fusion weight can reproduce dynamic fusion.

\begin{table}[htbp]
  \centering
  \caption{Fixed fusion gates versus learned FGDF. A dash indicates no successful rollout was available for the reported force statistic.}
  \label{tab:fixed_fusion_gates}
  \begin{tabular*}{\linewidth}{@{}l@{\extracolsep{\fill}}rrrr@{}}
    \toprule
    & \multicolumn{2}{c}{Gear Assembly} & \multicolumn{2}{c}{Box Flipping} \\
    \cmidrule(lr){2-3}\cmidrule(l){4-5}
    Fusion mechanism & SR $\uparrow$ & MaxF $\downarrow$ & SR $\uparrow$ & MaxF $\downarrow$ \\
    \midrule
    Fixed $\alpha=0.2$ & 80.0 & 8.2 & 55.0 & 21.3 \\
    Fixed $\alpha=0.5$ & 0.0 & --- & 40.0 & 21.5 \\
    Fixed $\alpha=0.8$ & 0.0 & --- & 0.0 & --- \\
    Learned FGDF & \textbf{90.0} & \textbf{7.6} & \textbf{95.0} & \textbf{10.9} \\
    \bottomrule
  \end{tabular*}
\end{table}

Learned FGDF substantially outperforms all constant gates (Tab.~\ref{tab:fixed_fusion_gates}). With $\alpha=0.2$, SR decreases to 80.0\% on Gear Assembly and 55.0\% on Box Flipping. Larger constant force weights cause more failures. Gear Assembly SR is zero for $\alpha=0.5$ and $0.8$, and both tasks fail with $\alpha=0.8$.
In Gear Assembly, large fixed weights can cause early gripper release during the non-contact phase. A possible explanation is that excessive force-feature weighting leads the policy to take the features as evidence that insertion is complete. The observed failures and layer-wise visualizations in Fig.~\ref{fig:fgdf_gate_visualization} support the need for fusion weights that vary across manipulation stages and AE layers.

\subsection{Qualitative results}

\textbf{Additional FAMI inference rate visualization.}
We provide additional visualizations of FAMI's predicted inference rate ratio on USB Insertion and Box Flipping in Fig.~\ref{fig:usb_freq_appendix} and Fig.~\ref{fig:flip_freq_appendix}, respectively. Each example shows the raw force measurements, computed force variance, and predicted inference rate ratio over time, illustrating how FAMI adjusts the AE update rate in response to changing contact dynamics across trials.

\begin{figure}[t]
    \centering
    \begin{subfigure}{0.48\linewidth}
        \centering
        \includegraphics[width=\linewidth]{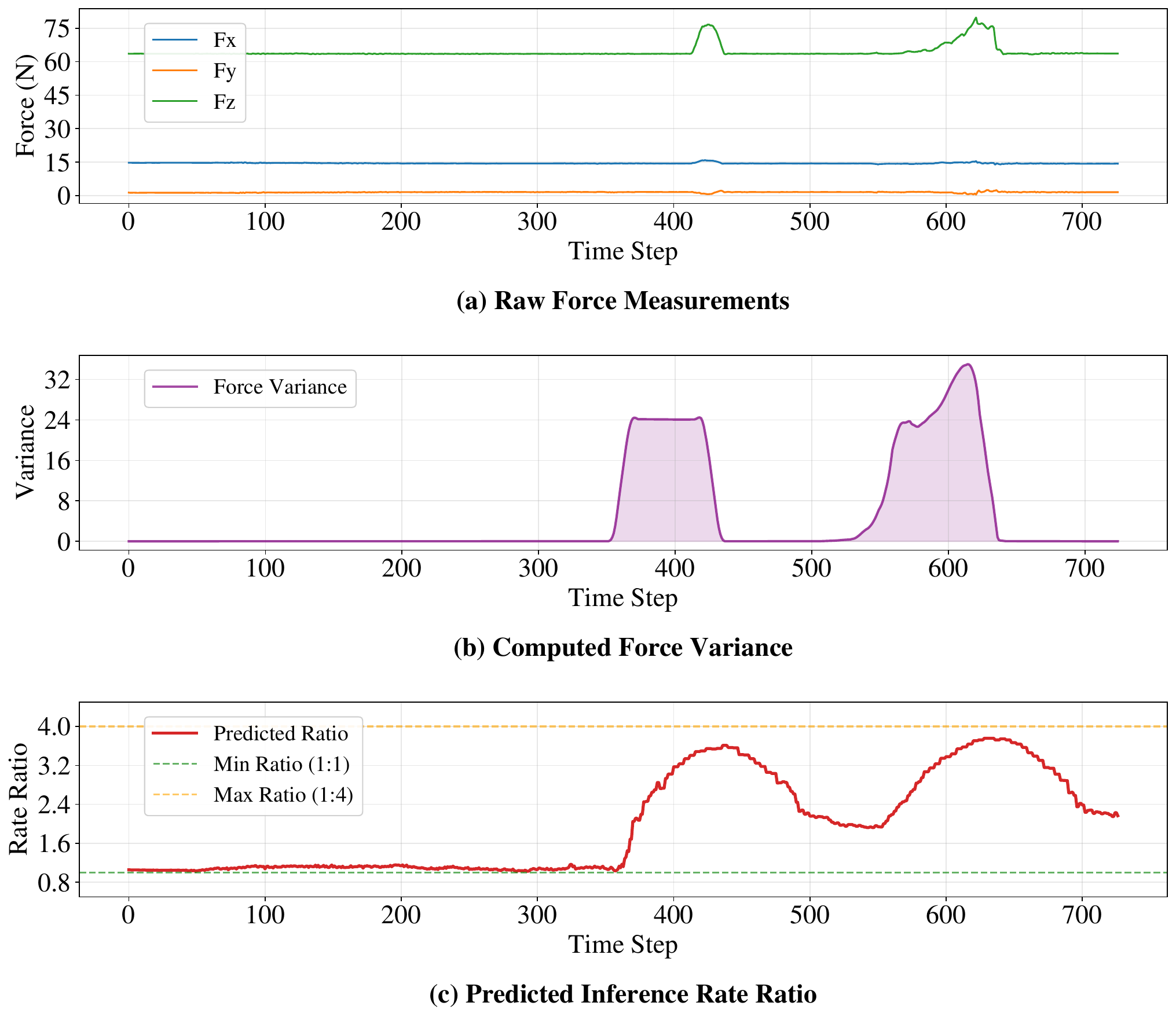}
        \caption{USB Insertion trial 1.}
    \end{subfigure}
    \hfill
    \begin{subfigure}{0.48\linewidth}
        \centering
        \includegraphics[width=\linewidth]{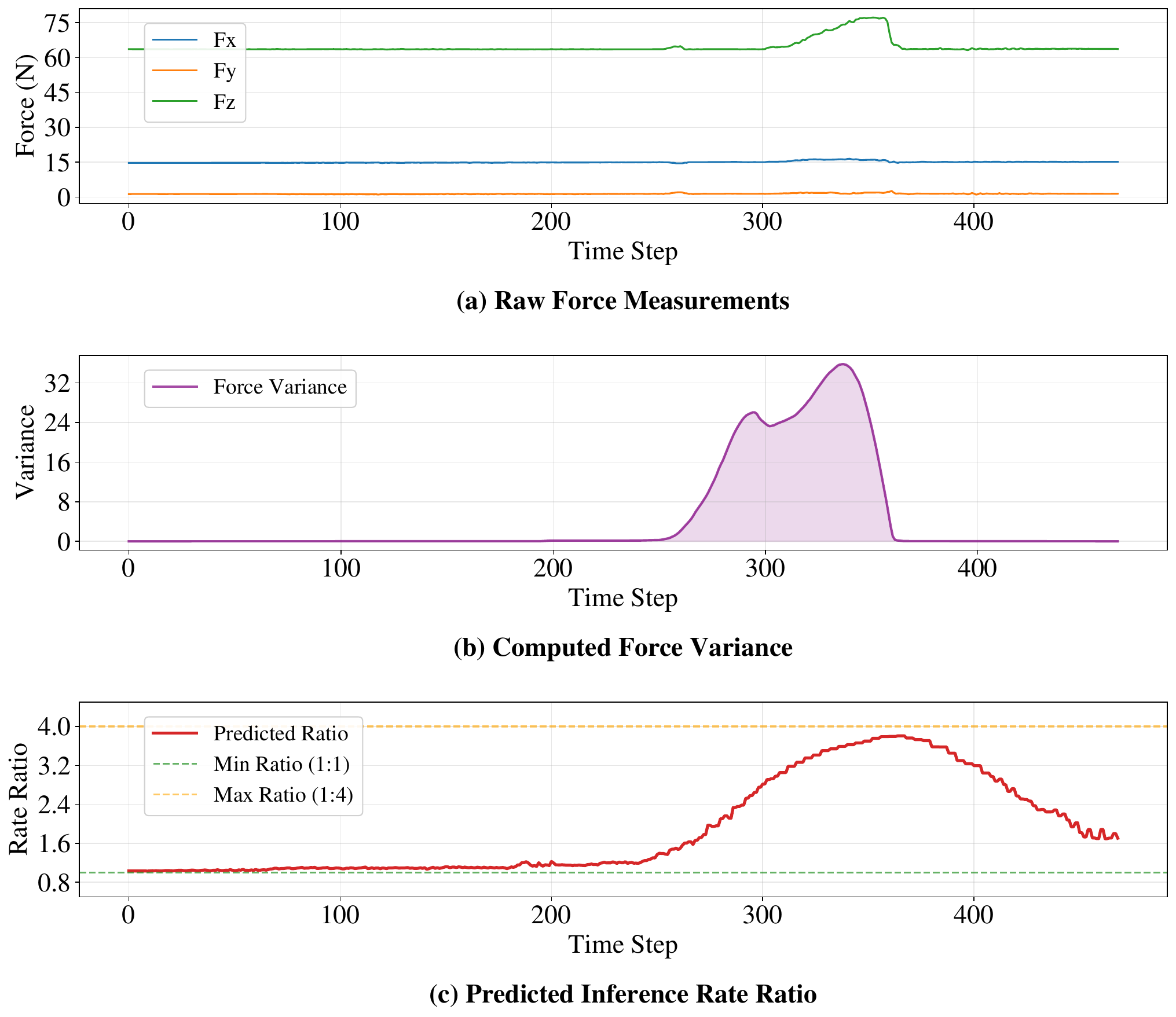}
        \caption{USB Insertion trial 2.}
    \end{subfigure}
    \caption{Additional visualization of FAMI's predicted inference rate ratio on the real-world \textbf{USB Insertion} task. Each trial shows raw force measurements, computed force variance, and the predicted VLM:AE inference rate ratio.}
    \label{fig:usb_freq_appendix}
\end{figure}

\begin{figure}[t]
    \centering
    \begin{subfigure}{0.48\linewidth}
        \centering
        \includegraphics[width=\linewidth]{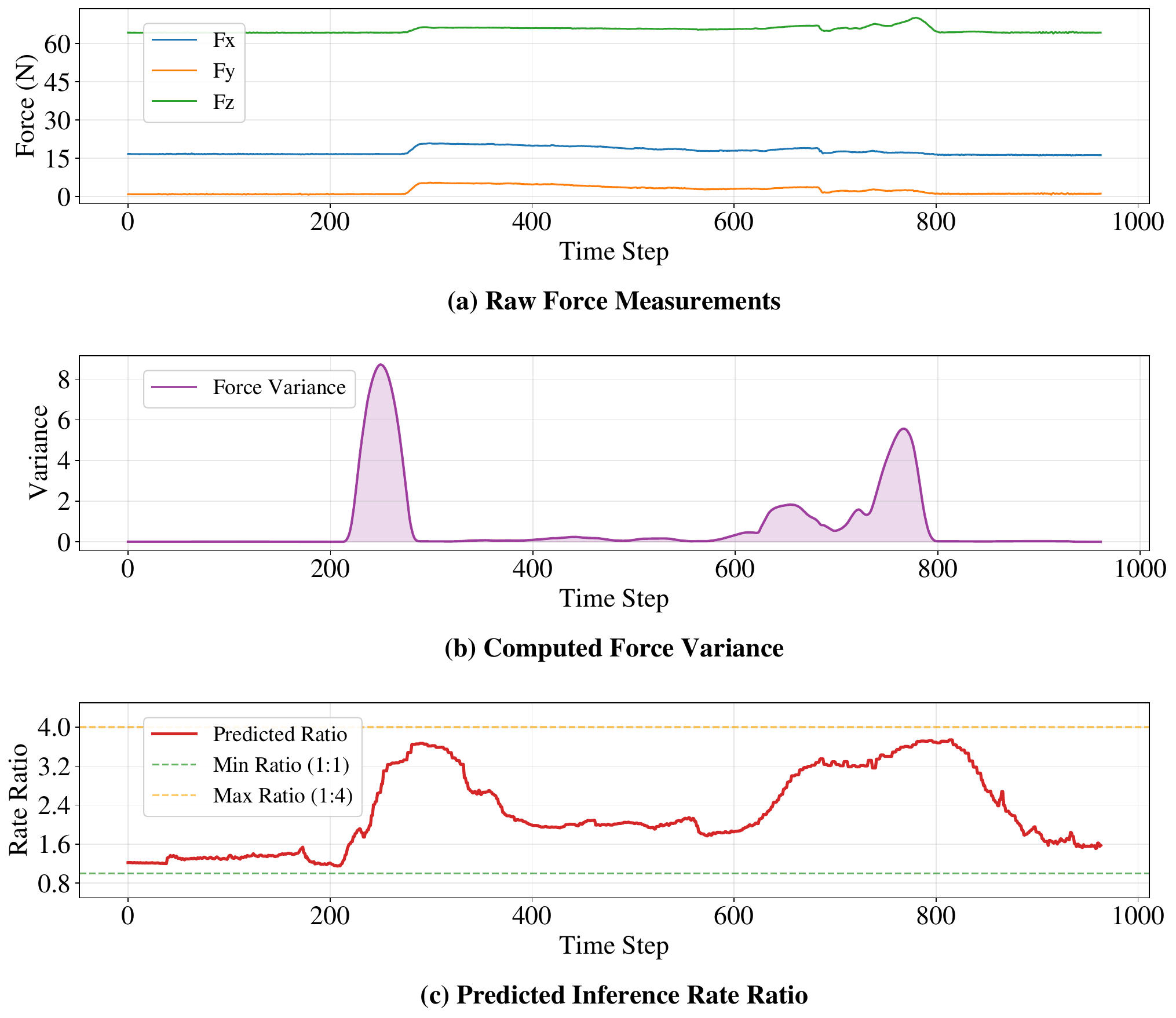}
        \caption{Box Flipping trial 1.}
    \end{subfigure}
    \hfill
    \begin{subfigure}{0.48\linewidth}
        \centering
        \includegraphics[width=\linewidth]{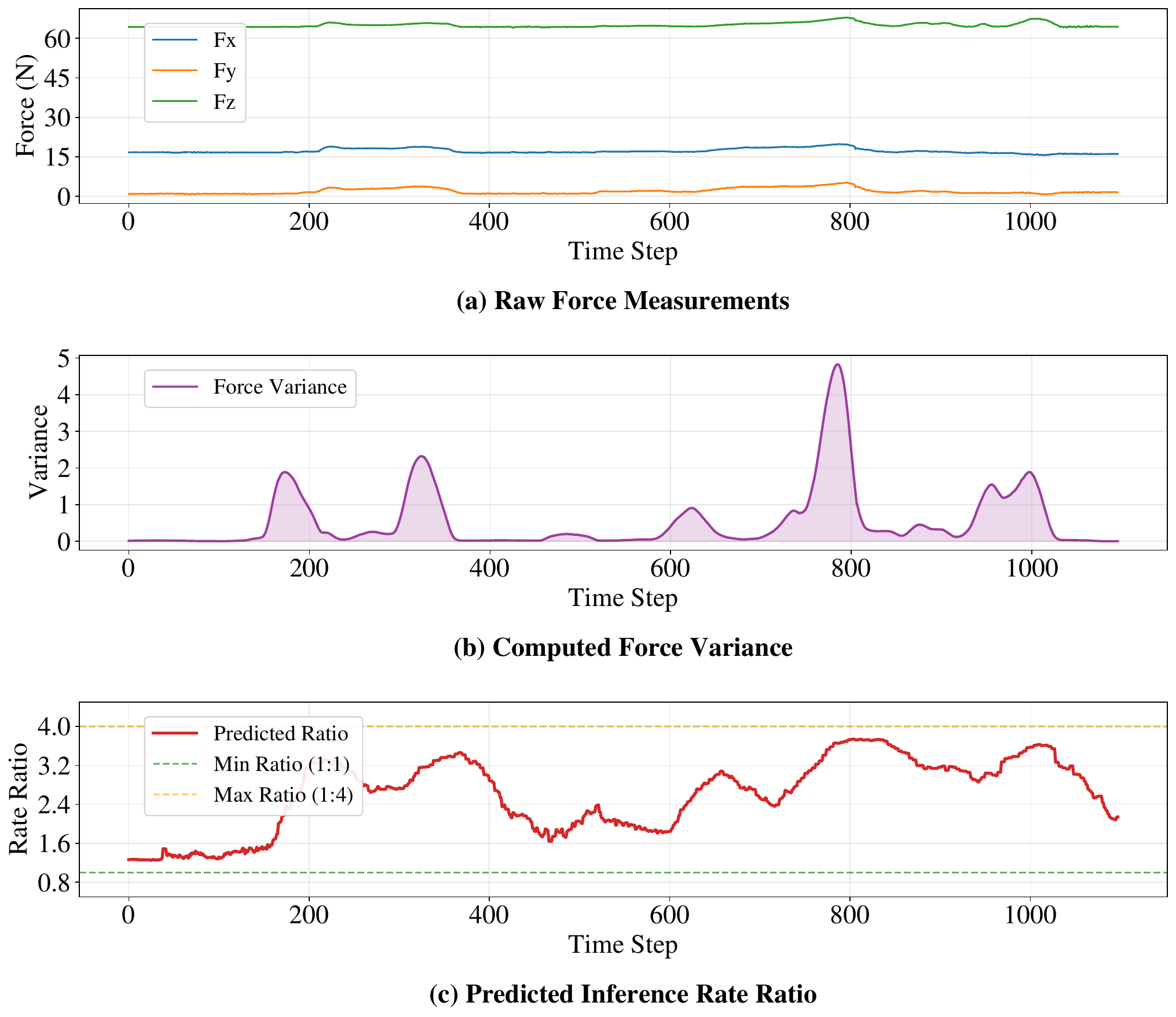}
        \caption{Box Flipping trial 2.}
    \end{subfigure}
    \caption{Additional visualization of FAMI's predicted inference rate ratio on the real-world \textbf{Box Flipping} task. }
    \label{fig:flip_freq_appendix}
\end{figure}

\textbf{Additional FGDF gate visualization.}
We provide additional FGDF gate visualizations in Fig.~\ref{fig:usb_gate_appendix} and Fig.~\ref{fig:flip_gate_appendix}. 
Across different trials, the gate values are selectively activated around force-sensitive phases, showing that FGDF dynamically adjusts the contribution of force-conditioned action features according to contact dynamics.

\begin{figure}[t]
    \centering
    \begin{subfigure}{0.48\linewidth}
        \centering
        \includegraphics[width=\linewidth]{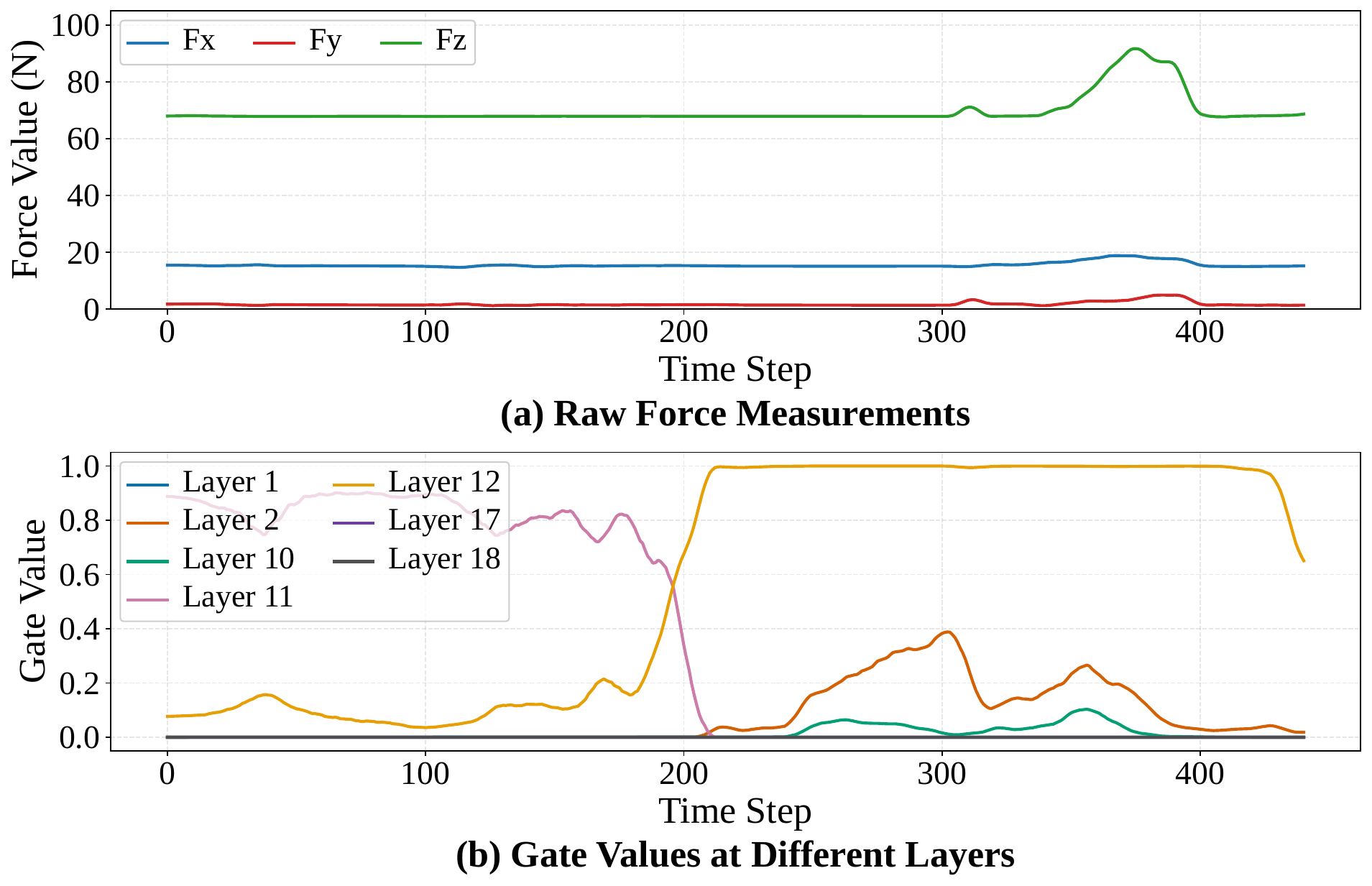}
        \caption{USB Insertion trial 1.}
    \end{subfigure}
    \hfill
    \begin{subfigure}{0.48\linewidth}
        \centering
        \includegraphics[width=\linewidth]{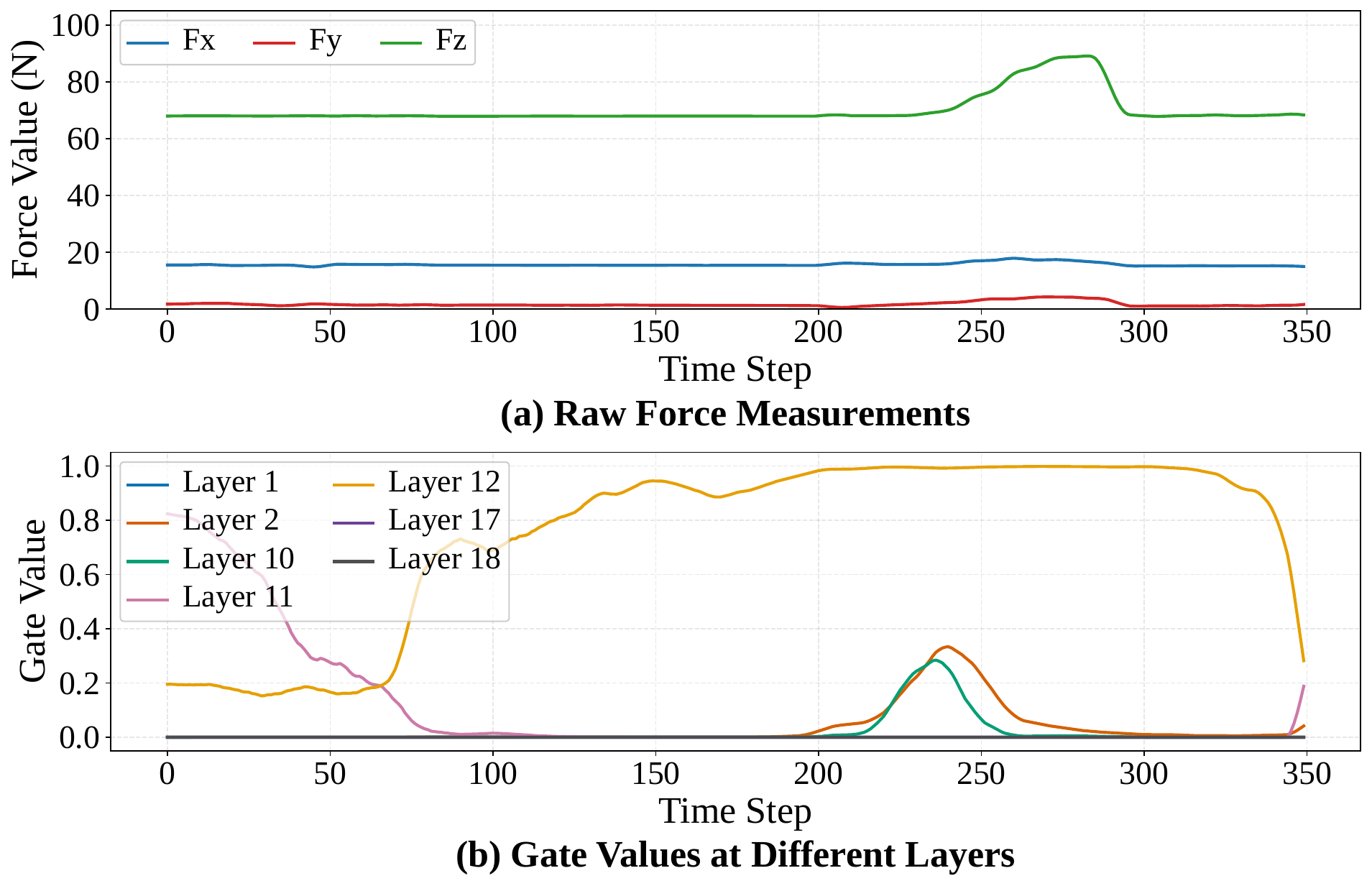}
        \caption{USB Insertion trial 2.}
    \end{subfigure}
    \caption{Additional visualization of FGDF gate values on the real-world \textbf{USB Insertion} task. The gate activations vary across time and layers, indicating adaptive force-feature fusion during contact-sensitive insertion phases.}
    \label{fig:usb_gate_appendix}
\end{figure}

\begin{figure}[t]
    \centering
    \begin{subfigure}{0.48\linewidth}
        \centering
        \includegraphics[width=\linewidth]{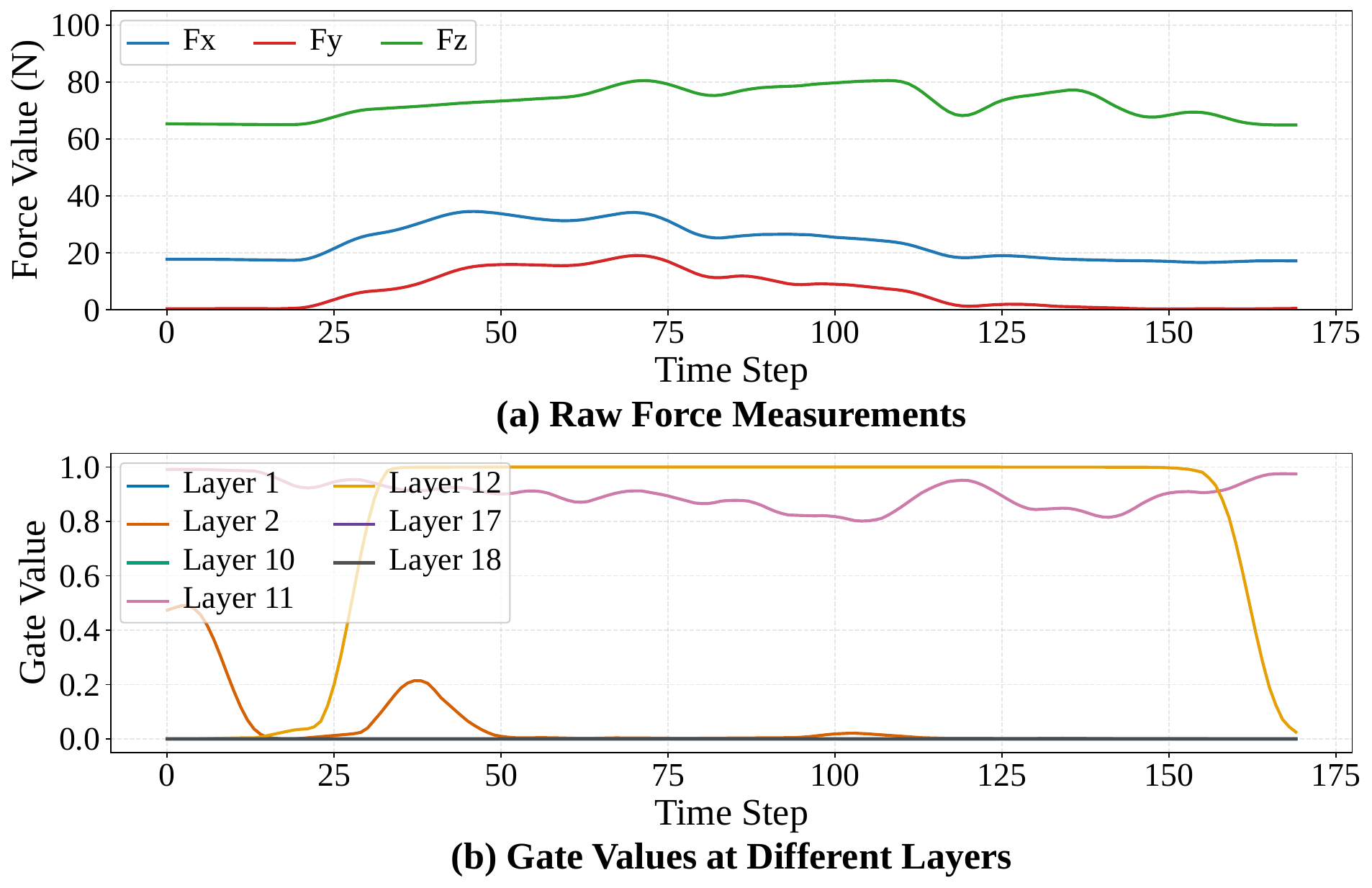}
        \caption{Box Flipping trial 1.}
    \end{subfigure}
    \hfill
    \begin{subfigure}{0.48\linewidth}
        \centering
        \includegraphics[width=\linewidth]{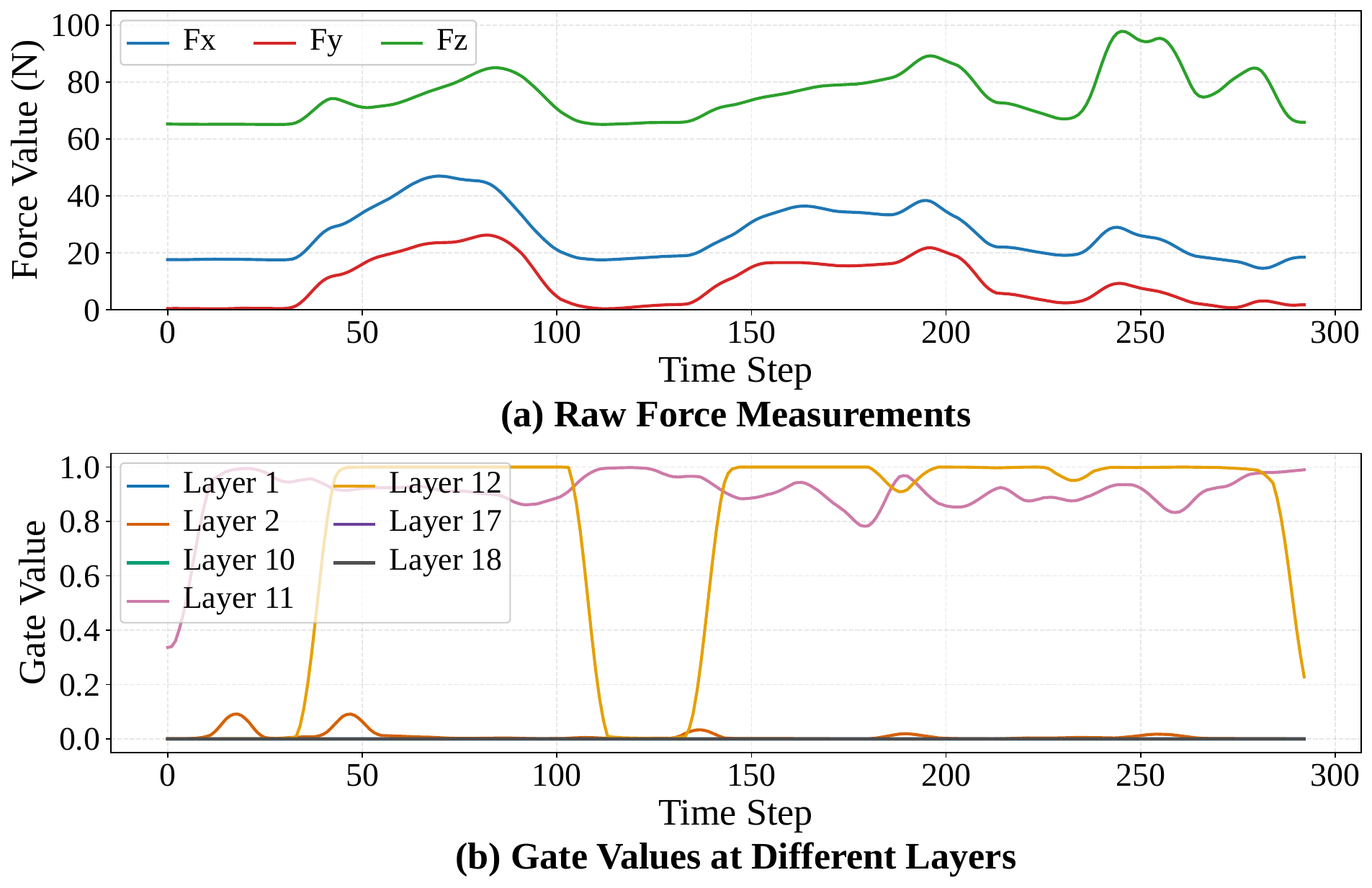}
        \caption{Box Flipping trial 2.}
    \end{subfigure}
    \caption{Additional visualization of FGDF gate values on the real-world \textbf{Box Flipping} task. The gate activations increase during force-sensitive contact phases, supporting dynamic visual-force fusion for sustained interaction.}
    \label{fig:flip_gate_appendix}
\end{figure}

\textbf{Real-world visualizations.}
As shown in Fig.~\ref{fig:task_execution_usb}, Fig.~\ref{fig:task_execution_gear}, Fig.~\ref{fig:task_execution_box}, Fig.~\ref{fig:task_execution_wiping}, and Fig.~\ref{fig:task_execution_peeling}, we visualize the execution process of the tasks in detail. Our observations show that our model can adapt to various perturbations, including angular rotation, unseen colors, and boxes of different shapes, with strong generalization ability and robustness.
For the USB Insertion task, the model can adapt to angularly rotated sockets and adjust its position via force feedback to achieve precise insertion even when misaligned.
For the Gear Assembly task, the model possesses strong generalization to unseen gears, including those with unseen colors that were not included in training, and can successfully insert such gears.
For the Box Flipping task, the model has excellent force control to prevent box slipping and features effective adaptation to various unseen box sizes.
For the Board Wiping task, the model can generalize to markers of different colors and text.
For the Cucumber Peeling task, the model maintains sustained contact between the peeler and cucumber surface and removes the skin in continuous strips.
Overall, our model demonstrates strong robustness and generalization ability in real-world manipulation tasks, which verifies its practical applicability and reliability.}

\textbf{Limitation analysis.}
As an imitation learning approach, FAVLA depends on the quality and diversity of demonstrations, and data collection currently requires a force-aware teleoperation system. 
Therefore, the model performance is limited by the quality of the collected demonstrations and the efficiency of the current data collection pipeline. 
In future work, we will improve the force-feedback-based data collection process to collect higher-quality demonstrations more efficiently, and further incorporate stronger task-completion and recovery mechanisms. Additionally, we will explore combining FAVLA with reinforcement learning to further improve the robustness of FAVLA.

\begin{figure}[htbp] 
    \centering
    
    \includegraphics[width=0.95\textwidth]{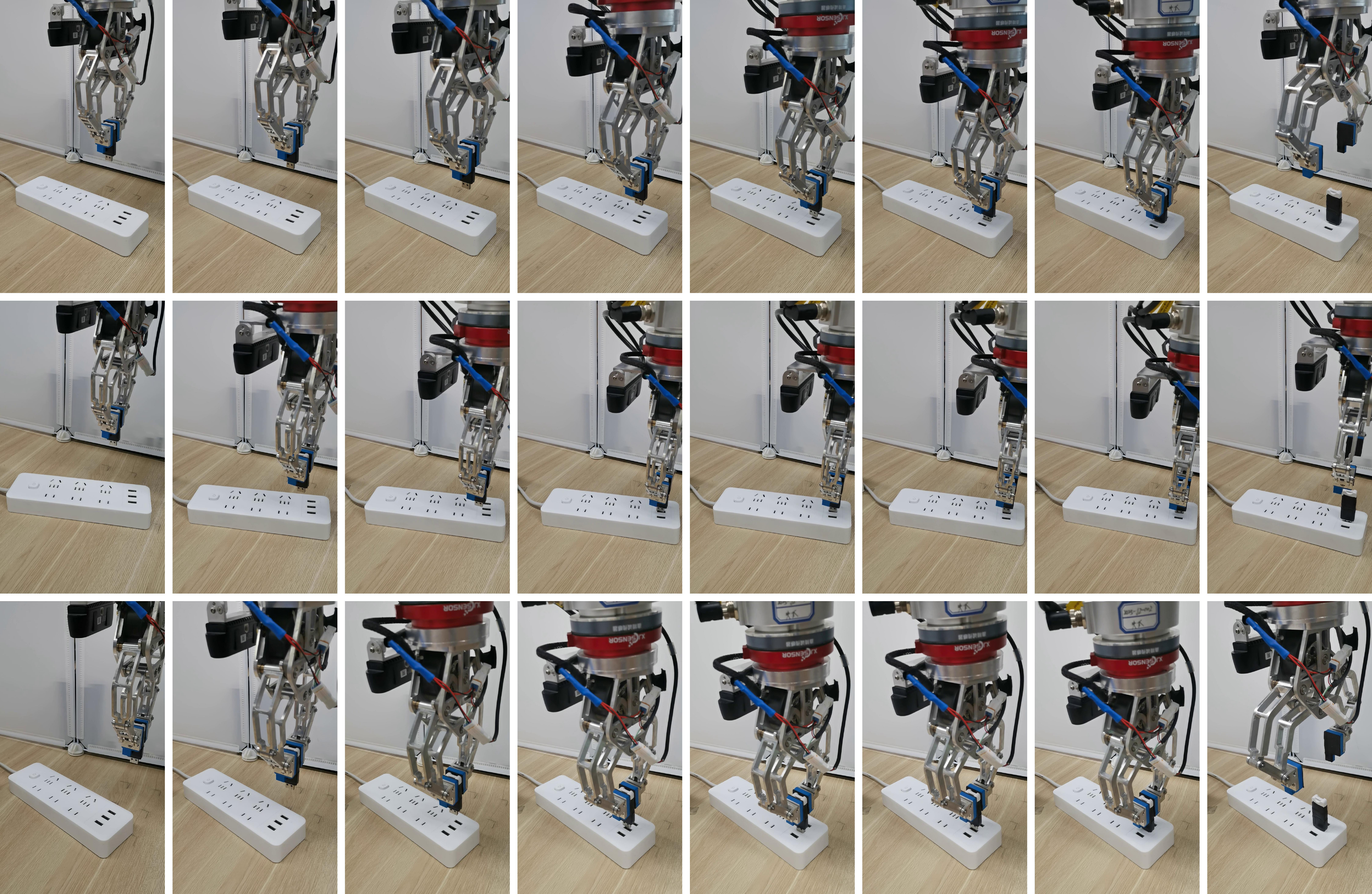} 
    \caption{Qualitative results on the real-world \textbf{USB Insertion} task. Our model successfully handles power strips with different rotations.}
    \label{fig:task_execution_usb}
    
    \vspace{8pt} 
    \includegraphics[width=0.95\textwidth]{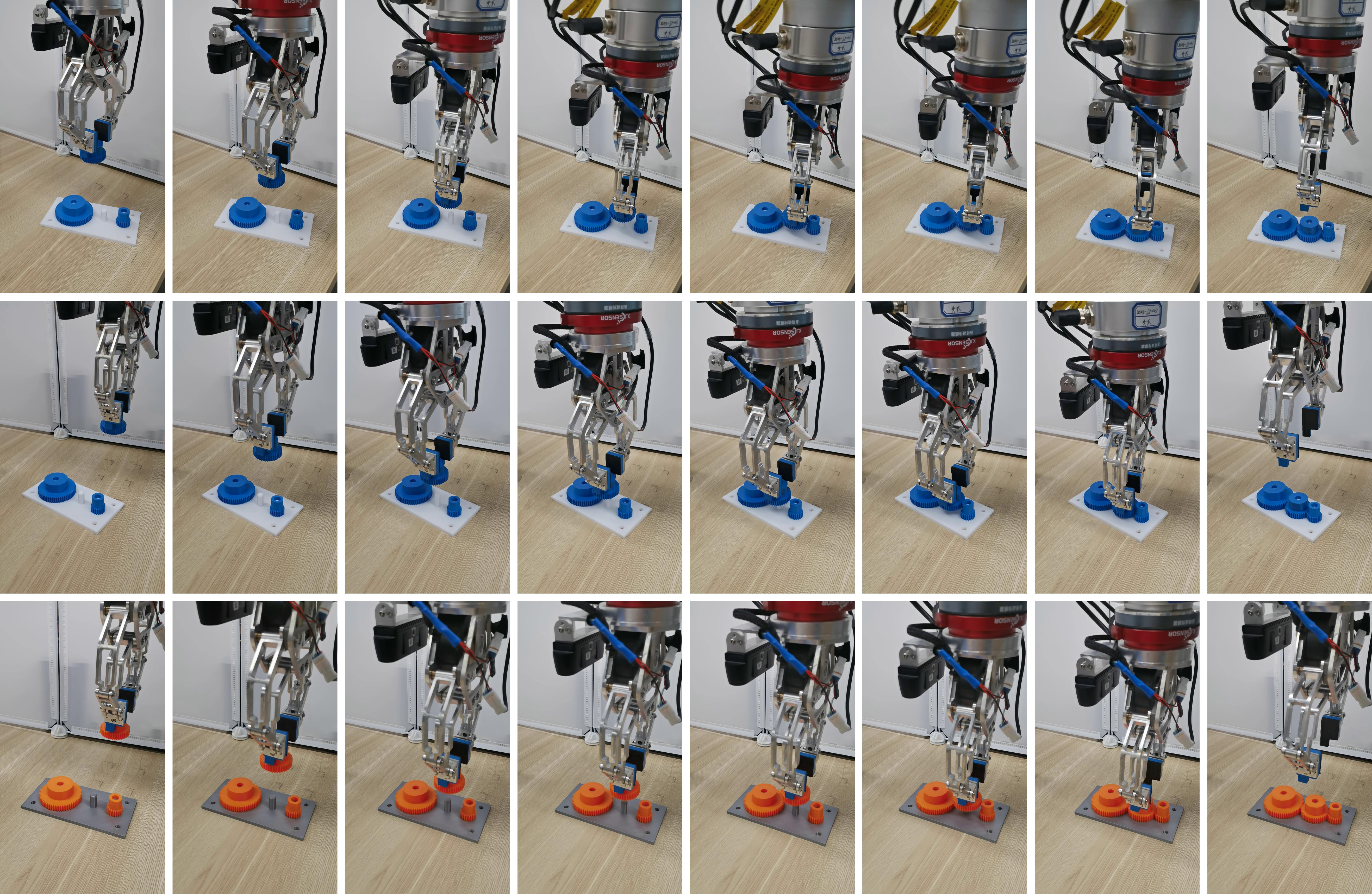} 
    \caption{Qualitative results on the real-world \textbf{Gear Assembly} task. Our model successfully generalizes to gears of unseen colors.}
    \label{fig:task_execution_gear}
\end{figure}

\begin{figure}[htbp]
    \centering
    \includegraphics[width=0.95\textwidth]{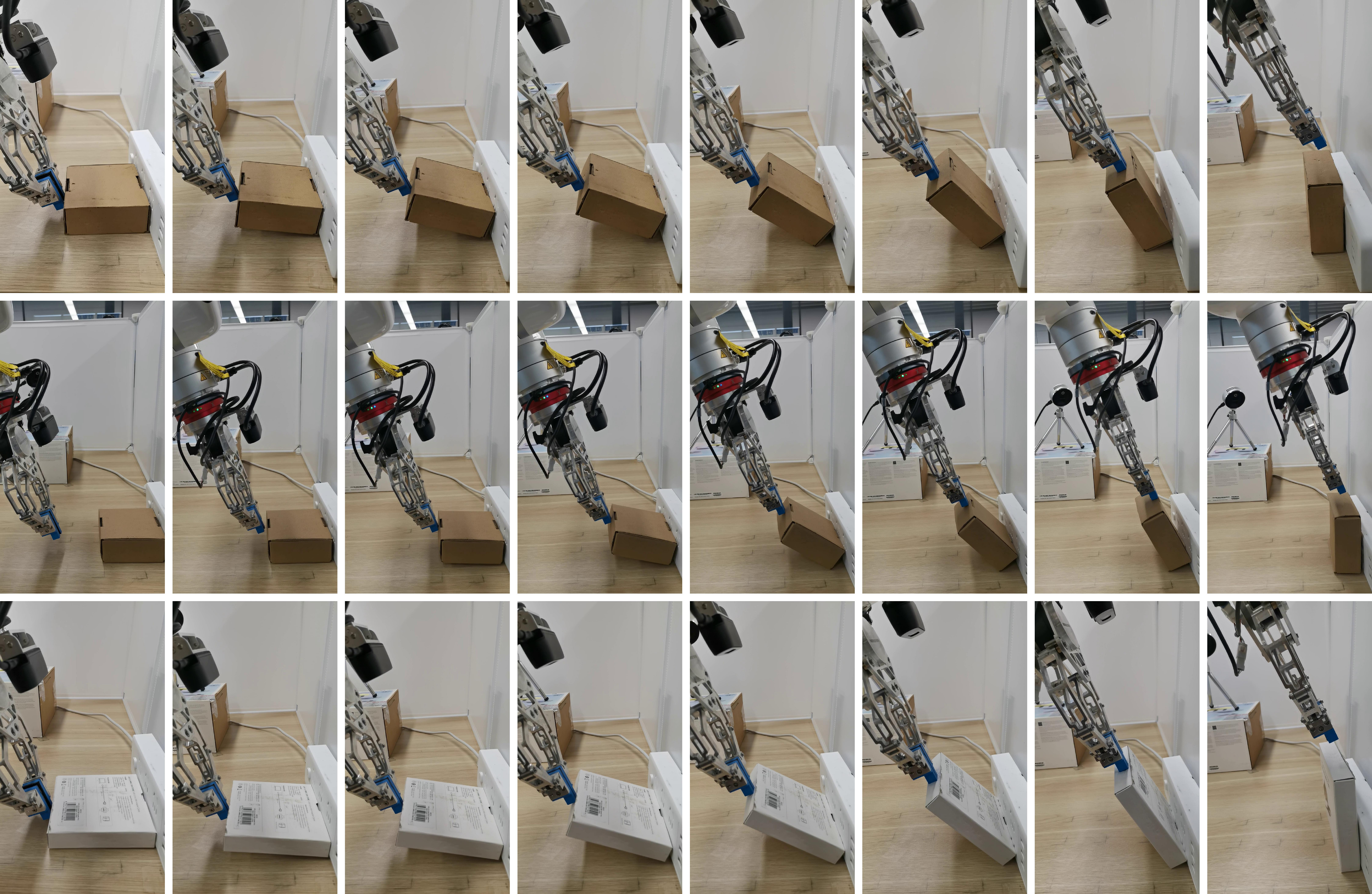} 
    \caption{Qualitative results on the real-world \textbf{Box Flipping} task. Our model generalizes to boxes of unseen sizes.}
    \label{fig:task_execution_box}
    \vspace{8pt} 
    \includegraphics[width=0.95\textwidth]{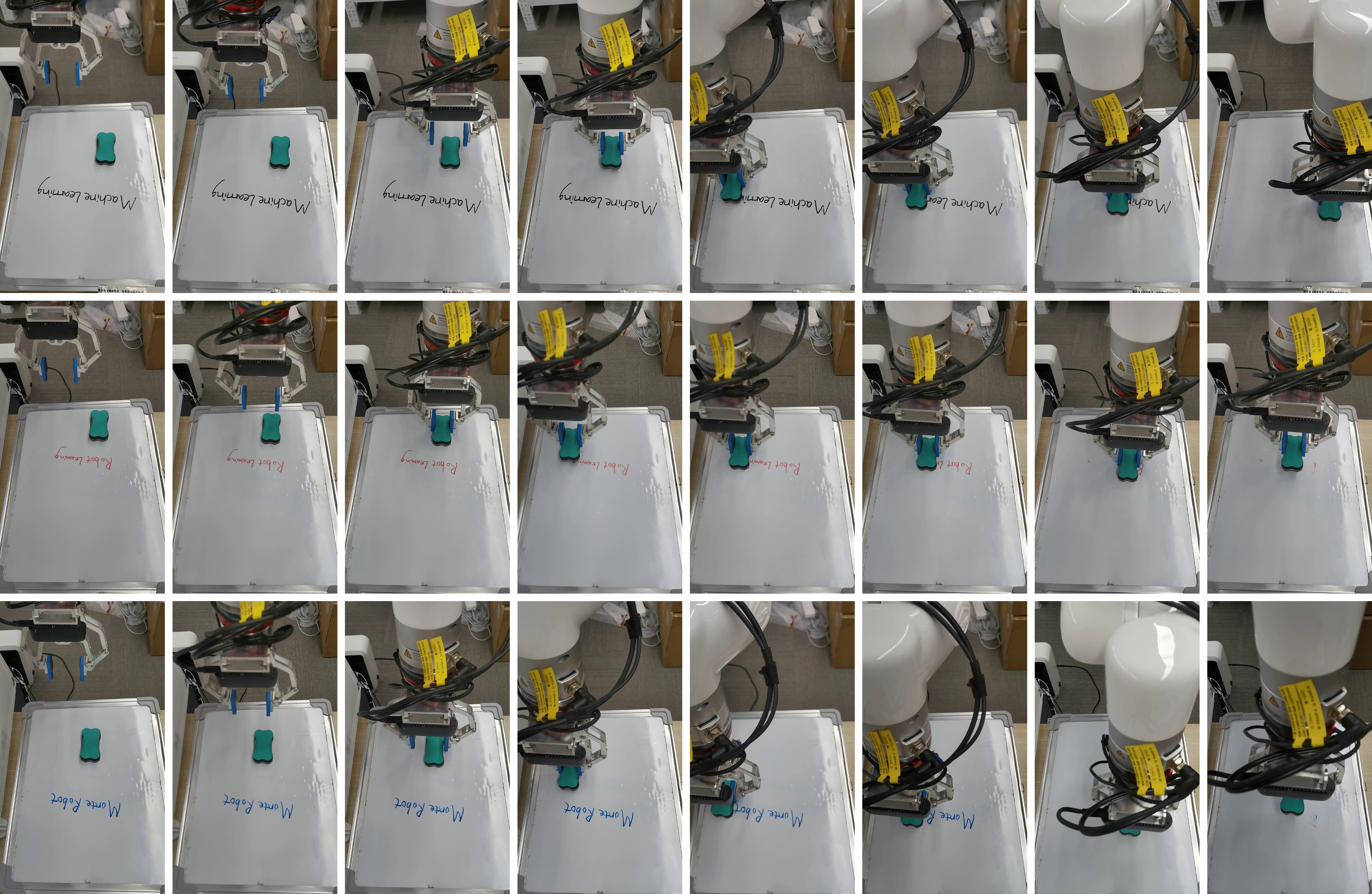} 
  \caption{Qualitative results on the real-world \textbf{Board Wiping} task. Our model generalizes to marker ink of unseen colors.}
    \label{fig:task_execution_wiping}
\end{figure}

\begin{figure}[tb]
    \centering
    \includegraphics[width=0.95\textwidth]{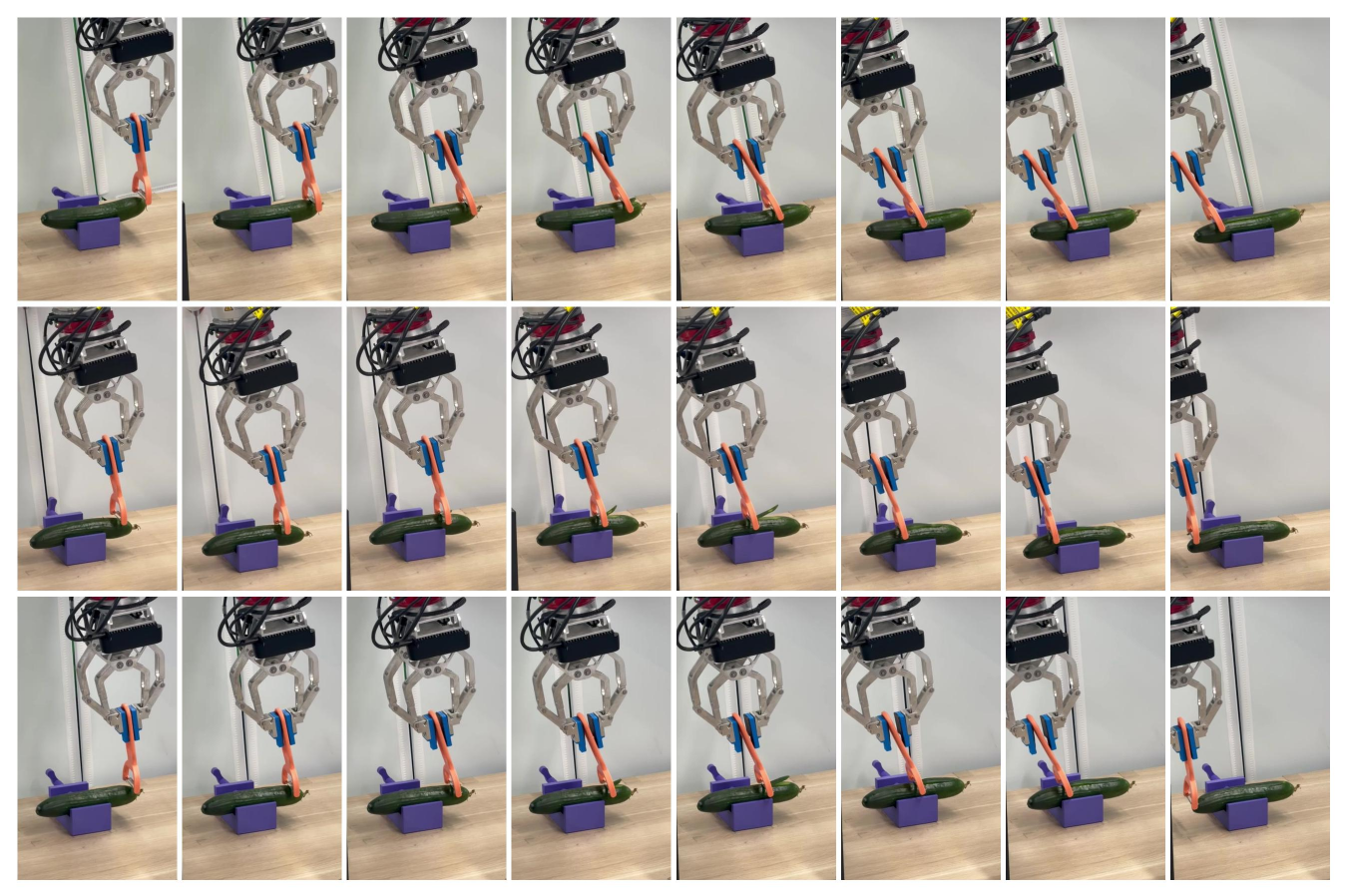}
    \caption{Qualitative results on the real-world \textbf{Cucumber Peeling} task. Our model maintains sustained contact between the peeler and cucumber surface and peels the skin into continuous strips.}
    \label{fig:task_execution_peeling}
\end{figure}

%% file: checklist.tex
\clearpage
\section*{NeurIPS Paper Checklist}

\begin{enumerate}

\item {\bf Claims}
    \item[] Question: Do the main claims made in the abstract and introduction accurately reflect the paper's contributions and scope?
    \item[] Answer: \answerYes{} 
    \item[] Justification: The abstract and introduction state the main contributions of FAVLA, including force-adaptive multi-rate inference, force-guided dynamic fusion, and real-robot evaluation on contact-rich manipulation tasks. 

\item {\bf Limitations}
    \item[] Question: Does the paper discuss the limitations of the work performed by the authors?
    \item[] Answer: \answerYes{} 
    \item[] Justification: The appendix discusses limitations of FAVLA, including its dependence on demonstration quality and diversity, the need for a force-aware teleoperation system, and future directions for improving data collection and performance.

\item {\bf Theory assumptions and proofs}
    \item[] Question: For each theoretical result, does the paper provide the full set of assumptions and a complete (and correct) proof?
    \item[] Answer: \answerNA{} 
    \item[] Justification: The paper involves a real-world robotic system and does not include theoretical results or formal proofs.

    \item {\bf Experimental result reproducibility}
    \item[] Question: Does the paper fully disclose all the information needed to reproduce the main experimental results of the paper to the extent that it affects the main claims and/or conclusions of the paper (regardless of whether the code and data are provided or not)?
    \item[] Answer: \answerYes{} 
    \item[] Justification: The paper describes the model architecture, training setup, datasets, task definitions, evaluation metrics, and implementation details in the main text and appendix.

\item {\bf Open access to data and code}
    \item[] Question: Does the paper provide open access to the data and code, with sufficient instructions to faithfully reproduce the main experimental results, as described in supplemental material?
    \item[] Answer: \answerNo{} 
    \item[] Justification: The paper provides detailed implementation and dataset descriptions. We plan to release the code publicly at our project website.

\item {\bf Experimental setting/details}
    \item[] Question: Does the paper specify all the training and test details (e.g., data splits, hyperparameters, how they were chosen, type of optimizer) necessary to understand the results?
    \item[] Answer: \answerYes{} 
    \item[] Justification: The experiments section and appendix report the tasks, trial numbers, model settings, optimizer, learning rate, batch size, training iterations, and inference settings.

\item {\bf Experiment statistical significance}
    \item[] Question: Does the paper report error bars suitably and correctly defined or other appropriate information about the statistical significance of the experiments?
    \item[] Answer: \answerNo{} 
    \item[] Justification: The paper reports success rates and average peak forces over real-robot trials, but does not include formal error bars or statistical significance tests due to the cost of real-robot evaluation.

\item {\bf Experiments compute resources}
    \item[] Question: For each experiment, does the paper provide sufficient information on the computer resources (type of compute workers, memory, time of execution) needed to reproduce the experiments?
    \item[] Answer: \answerYes{} 
    \item[] Justification: The appendix reports the training GPU, training time, and inference latency measured on an NVIDIA RTX 4090 GPU.
    
\item {\bf Code of ethics}
    \item[] Question: Does the research conducted in the paper conform, in every respect, with the NeurIPS Code of Ethics \url{https://neurips.cc/public/EthicsGuidelines}?
    \item[] Answer: \answerYes{} 
    \item[] Justification: The work is conducted on controlled real-robot manipulation tasks and does not involve sensitive personal data or deceptive practices.

\item {\bf Broader impacts}
    \item[] Question: Does the paper discuss both potential positive societal impacts and negative societal impacts of the work performed?
    \item[] Answer: \answerNo{} 
    \item[] Justification: The paper focuses on technical contributions for robotic manipulation and does not include a detailed broader impact discussion.
    
\item {\bf Safeguards}
    \item[] Question: Does the paper describe safeguards that have been put in place for responsible release of data or models that have a high risk for misuse (e.g., pre-trained language models, image generators, or scraped datasets)?
    \item[] Answer: \answerNA{} 
    \item[] Justification: The paper does not release high-risk generative models or scraped datasets requiring special safeguards.

\item {\bf Licenses for existing assets}
    \item[] Question: Are the creators or original owners of assets (e.g., code, data, models), used in the paper, properly credited and are the license and terms of use explicitly mentioned and properly respected?
    \item[] Answer: \answerYes{} 
    \item[] Justification: Existing models, baselines, and libraries used in the paper are cited in the related work and experimental setup.

\item {\bf New assets}
    \item[] Question: Are new assets introduced in the paper well documented and is the documentation provided alongside the assets?
    \item[] Answer: \answerYes{} 
    \item[] Justification: The collected real-robot dataset is documented in the appendix with task descriptions, dataset statistics, observation/action spaces, and data collection details.

\item {\bf Crowdsourcing and research with human subjects}
    \item[] Question: For crowdsourcing experiments and research with human subjects, does the paper include the full text of instructions given to participants and screenshots, if applicable, as well as details about compensation (if any)? 
    \item[] Answer: \answerNA{} 
    \item[] Justification: The paper does not involve crowdsourcing or human-subject studies. A human operator only teleoperates the robot for demonstration collection.

\item {\bf Institutional review board (IRB) approvals or equivalent for research with human subjects}
    \item[] Question: Does the paper describe potential risks incurred by study participants, whether such risks were disclosed to the subjects, and whether Institutional Review Board (IRB) approvals (or an equivalent approval/review based on the requirements of your country or institution) were obtained?
    \item[] Answer: \answerNA{} 
    \item[] Justification: The work does not conduct human-subject experiments or collect personal data from participants.

\item {\bf Declaration of LLM usage}
    \item[] Question: Does the paper describe the usage of LLMs if it is an important, original, or non-standard component of the core methods in this research? Note that if the LLM is used only for writing, editing, or formatting purposes and does \emph{not} impact the core methodology, scientific rigor, or originality of the research, declaration is not required.
    \item[] Answer: \answerNA{} 
    \item[] Justification: LLMs are only used for grammar checking and wording refinement, and are not involved in the core method design, experiments, or scientific conclusions.

\end{enumerate}